\documentclass[letterpaper, 10 pt, conference]{ieeeconf}  

\IEEEoverridecommandlockouts  
\usepackage{textcomp}

\usepackage{sansmath}
\usepackage{graphicx}
\usepackage{scalerel}
\usepackage{amsmath}
\usepackage{xspace}
\usepackage{multirow}
\usepackage{diagbox}
\usepackage{amsfonts}
\usepackage{mathtools}
\usepackage[citecolor=magenta, colorlinks=true]{hyperref}
\usepackage[english]{babel}
\usepackage[version=4]{mhchem}
\usepackage{siunitx}
\usepackage{blindtext}
\usepackage{longtable,tabularx}
\usepackage[utf8]{inputenc}
\usepackage{graphicx}
\usepackage{subcaption}
\usepackage[font=small]{caption}
\usepackage[version=4]{mhchem}
\usepackage{longtable,tabularx}
\usepackage{algorithm}
\usepackage{algpseudocode}
\usepackage{float}
\usepackage{titlesec}
\usepackage{tikz}
\usepackage{pgfplots}
\pgfplotsset{compat=1.17}
\usepackage[compress]{cite}
\usepackage{booktabs}
\usepackage{comment}
\usepackage{pgfplotstable} 
\usetikzlibrary{positioning, shapes.geometric, fit}
\usetikzlibrary{calc}

\newcommand{\ours}{\text{MA-ICBF} }
\newcommand{\macbf}{\text{MA-CBF}}
\newcommand{\gcbf}{\text{GCBF}}

\newcommand{\bx}{{\boldsymbol{x}}}
\newcommand{\bu}{{\boldsymbol{u}}}
\newcommand{\by}{{\boldsymbol{y}}}

\newcommand{\bv}{{\boldsymbol{v}}}
\newcommand{\bp}{{\boldsymbol{p}}}
\newcommand{\scaletikz}{1}
\newcommand{\agentempty}{\textbf{Number of Agents }}
\newcommand{\agentmaze}{\textbf{Number of Agents }}

\newcommand{\widthtikz}{8.5cm}
\newcommand{\heighttikz}{5.5cm}

\newtheorem{theorem}{\bf Theorem}

\newtheorem{problem}{\bf Problem}
\newtheorem{definition}{\bf Definition}

\usepackage{dirtytalk}

\usepackage{float}

\title{\LARGE \bf
Decentralized Safe and Scalable Multi-Agent Control under Limited Actuation 
}

\author{Vrushabh Zinage$^{1}$, Abhishek Jha $^{2}$,  
Rohan Chandra $^{3}$, Efstathios Bakolas $^{1}$
\thanks{$^{1}$Vrushabh Zinage and Efstathios Bakolas are with the Department of Aerospace Engineering and Engineering Mechanics, University of Texas at Austin
        {\tt\small vrushabh.zinage@utexas.edu, bakolas@austin.utexas.edu}}%
        \thanks{$^{2}$ Abhishek Jha is with the Department of Mechanical Engineering, Delhi Technological University, New Delhi, India
        {\tt\small abhishekjha\_me20a8\_31@dtu.ac.in}}
        \thanks{$^{3}$ Rohan Chandra is with the Department of Computer Science at the University of Virginia, USA
        {\tt\small rohanchandra@virginia.edu }}
}
\begin{document}

\bibliographystyle{IEEEtran}

\maketitle
\thispagestyle{empty}
\pagestyle{empty}

\begin{abstract}
To deploy safe and agile robots in cluttered environments, there is a need to develop fully decentralized controllers that guarantee safety, respect actuation limits, prevent deadlocks, and scale to thousands of agents. Current approaches fall short of meeting all these goals: optimization-based methods ensure safety but lack scalability, while learning-based methods scale but do not guarantee safety. We propose a novel algorithm to achieve safe and scalable control for multiple agents under limited actuation. Specifically, our approach includes: $(i)$ learning a decentralized neural Integral Control Barrier function (neural ICBF) for scalable, input-constrained control, $(ii)$ embedding a lightweight decentralized Model Predictive Control-based Integral Control Barrier Function (MPC-ICBF) into the neural network policy to ensure safety while maintaining scalability, and $(iii)$ introducing a novel method to minimize deadlocks based on gradient-based optimization techniques from machine learning to address local minima in deadlocks. Our numerical simulations show that this approach outperforms state-of-the-art multi-agent control algorithms in terms of safety, input constraint satisfaction, and minimizing deadlocks. Additionally, we demonstrate strong generalization across scenarios with varying agent counts, scaling up to 1000 agents. Videos and code are available at \href{https://maicbf.github.io/}{\textbf{https://maicbf.github.io/}}.
\end{abstract}

\section{Introduction}
We consider the problem of safe decentralized real-time multi-agent control in the presence of obstacles for a large number of agents under limited actuation. Such controllers are particularly important for robots that have to be deployed in cluttered environments with large number of agents such as in warehouse automation \cite{li2023motion_warehouse_automation}, autonomous driving \cite{claussmann2019review_autonomous_driving}, quadrotor swarms \cite{batra2022decentralized_quad_swarm_1,park2022online_quad_swarm_2}, and automated intersection management \cite{au2010motion_intersection_management} where the agents have to operate in real-time and remain safe at all times
Furthermore, in these applications, real physical robots have limited actuation capabilities and cannot exceed certain control limits. Therefore, it becomes critical for the designed safe decentralized controller to satisfy input constraints to ensure the feasible deployment of robots. Lastly, as the environment gets more cluttered, resolving and preventing deadlocks becomes necessary as well.

There is a vast literature on this topic. Characteristic examples include references \cite{sharon2015conflict_classical_cbs_1,kottinger2022conflict_classical_cbs_2,dergachev2021distributed_classical_3,alonso2013optimal_classical_4, chandra2023decentralized,yu2022surprising_ma_lit_1,ames2019control_ma_lit_2,glotfelter2017nonsmooth_ma_lit_3,jankovic2021collision_ma_lit_4,cheng2020safe_ma_lit_5,garg2021robust_ma_lit_6,kohler2024distributed_opt_1,saccani2023model_opt_2,chen2020guaranteed_opt_3,wang2017safety_opt_4} which present either centralized or decentralized methods which have been popular choices due to their ability to guarantee safety, generalize to almost any nonlinear system and operate under limited actuation. Recently, control barrier functions (CBFs)~\cite{ames2019control_ma_lit_2} have been used increasingly used for guaranteed safety guarantees for nonlinear systems. For instance, CBF-based controllers have guaranteed safety for systems with higher relative degree \cite{xiao2019control_cbf_high_1,wang2021learning_cbf_high_2}, hybrid systems \cite{marley2024hybrid_1,lindemann2021learning_hybrid_2}, sampled data systems \cite{taylor2022safety_sampled_1,niu2021safety_sampled_2}, unknown nonlinear systems \cite{zinage2023neural_unknown_zinage_1,zinage2023neural_icbf,zinage2023neuralunknown}, input delay systems \cite{jankovic2018control_input_delay_systems}, non-control affine systems \cite{ames2020integral_ames,zinage2023neural_icbf,zinage2023disturbance_integral_zinage} and systems with additive bounded disturbances. However, the main challenge in extending these approaches to multi-agent systems is that they are not scalable and deadlock-free in general. Furthermore, synthesizing a CBF for multi-agent scenarios is non-trivial in general. Learning-based methods \cite{long2018towards_learning_lit_1,chen2017decentralized_learning_lit_2,everett2021collision_learning_lit_3,kamenev2022predictionnet_learning_lit_4,liu2020pic} including multi-agent reinforcement learning (MARL) \cite{zhang2021multi_marl_review, wang2022model_based_marl}, on the other hand, have demonstrated better scalability, with some methods scaling as much as up to $32$ agents~\cite{liu2020pic,yu2022surprising_mappo,nayak22informarl}. However, learning-based methods impose soft constraints (collision checking) on safety, as opposed to provable safety constraints imposed by CBFs, to generate collision-free trajectories, albeit without formal guarantees on safety, input constraint satisfaction, or deadlock resolution capabilities.
Furthermore, some learning-based algorithms do not exploit the fact that governing nonlinear system equations are known either apriori or via system identification methods \cite{nayak22informarl}. Consequently, these methods require large datasets comprising state-action pairs to train these MARL algorithms efficiently so that they can be competitive to control-theoretic approaches, which is impractical in many challenging real world scenarios.

\begin{figure}[t]
    \centering
   \begin{subfigure}[h]{.49\columnwidth}
    \includegraphics[width=\textwidth]{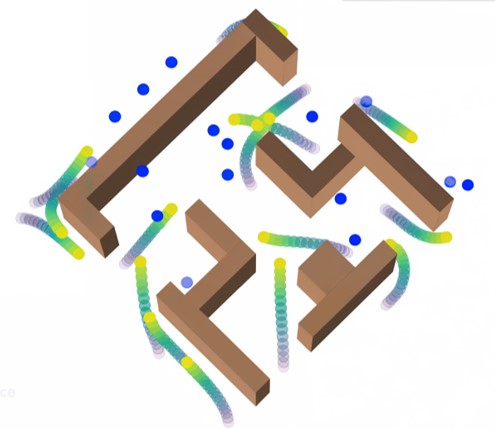}
    \caption{\texttt{Complex Maze}}
    \label{fig: lines_ped}
  \end{subfigure}
   \begin{subfigure}[h]{.49\columnwidth}
    \includegraphics[width=\textwidth]{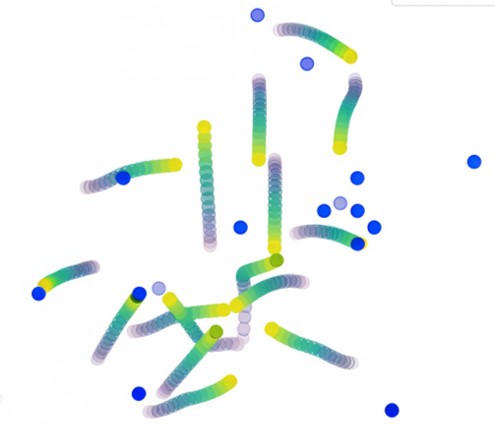}
    \caption{\texttt{Empty environment}}
    \label{fig: lines_veh}
  \end{subfigure}
    \caption{Demonstrating our decentralized safe and scalable multi-agent control in two environments. Blue and yellow denote goals and current positions, respectively.}
    \label{fig: cover}
    \vspace{-15pt}
\end{figure}

\begin{figure*}[h!]
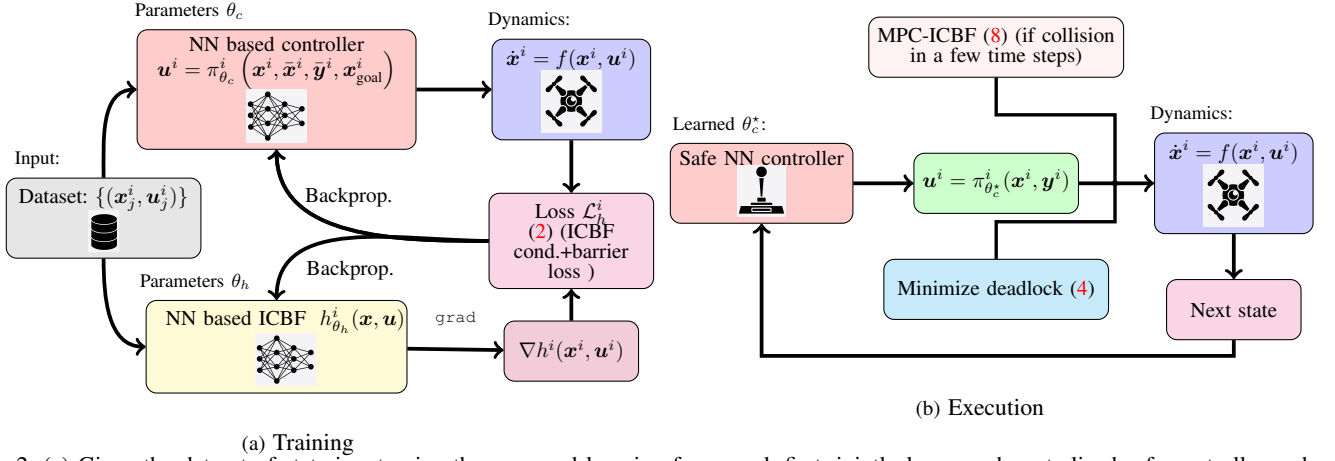



    \centering
    \begin{minipage}{0.49\textwidth}
        \centering
        \input{tikz/learning_approach_new}
        \subcaption{\small Training}
        \label{fig:learning_approach}
    \end{minipage}
    \hfill
    \begin{minipage}{0.49\textwidth}
        \centering
        \input{tikz/proposed_approach}
        \subcaption{\small Execution}
        \label{fig:proposed_approach}
    \end{minipage}
\caption{\small (a) Given the dataset of state-input pairs, the proposed learning framework first, jointly learns a decentralized safe controller and a decentralized ICBF for each agent $i\in V$. (b) Next, during implementation, \ours utilizes the learned safe decentralized controller to steer the agent to its goal position while at the same time guaranteeing safety (including input constraint satisfaction) and deadlock minimization via the MPC-ICBF \eqref{eqn:mpc_proposed_via_icbf} and the deadlock minimization modules \eqref{eqn:deadlock_resolve} respectively.
}
    \label{fig:proposed_approach_total}
    \vspace{-10pt}
\end{figure*}
Recently, a line of research~\cite{qin2021learning_macbf,zhang2023neural_gcbf} has combined the best of both classical optimization-based methods and learning-based methods to synthesize decentralized controllers that are scalable up to thousands of agents. The main idea of these approaches is to simultaneously learn safe feedback controllers and decentralized CBFs based neural certificates for each agent that ensure scalability and safety during planning.
Although these approaches provide probabilistic guarantees on collision avoidance, they do not guarantee safety \cite{qin2021learning_macbf,zhang2023neural_gcbf}, suffer from the problem of deadlock among these agents \cite{qin2021learning_macbf}, cannot operate under limited actuation \cite{zhang2023neural_gcbf}, and are restricted to control affine nonlinear systems \cite{ames2019control_ma_lit_2,tee2009barrier_cbf_1,zhang2023neural_gcbf}. Towards the goal of synthesizing decentralized controllers that are scalable and guarantee safety, our contributions are as follows:
\begin{enumerate}
    \item \textit{Safety and Scalability:} We propose a novel learning-based framework termed \ours based on Integral Control Barrier Functions (ICBFs) \cite{zinage2023neural_icbf,ames2020integral_ames} for safe multi-agent control that are scalable and are able to handle general nonlinear systems with input constraints.
    \item \textit{Limited Actuation:} As the proposed learning-based framework can only provide probabilistic guarantees, we embed a novel computationally light model predictive control (MPC) optimization-based framework termed MPC-ICBF into our approach that guarantees safety as well as input constraint satisfaction without compromising the overall computational time for our approach.
    \item \textit{Deadlock Resolution:} By drawing an analogy between well-established gradient-based methods for minimizing non-convex loss functions in ML and the problem of non-convexity (or the problem of local minima) that are common in deadlocks, we present a novel deadlock resolution technique to minimize deadlocks.
\end{enumerate}
The overall structure of the paper is as follows. 
Section \ref{sec:prelim_and_problem} discusses the preliminaries and the problem statement. Section \ref{sec:proposed_framework} discusses the proposed approach followed by results in Section \ref{sec:results}. Finally, we conclude in Section \ref{sec:conclusion}.
\section{Preliminaries and Problem Statement\label{sec:prelim_and_problem}}
\subsection{Nomenclature}
For integers $a$ and $b (\geq a)$, the set $[a;b]_d$ denotes the set of integers $\{a,a+1,\dots, b\}$. The superscript $i$ on any variable indicates the variable for the $i^{\text{th}}$ agent. The set $V = [1;N]_d$ denotes the set of indices for $N$ agents. For a set $\mathcal{A}$, $\partial\mathcal{A}$ denotes its boundary. 

\subsection{Preliminaries and assumptions}
We formulate the decentralized multi-agent control problem by $\langle N,\;M,\;\mathcal{X},\;\mathcal{X}_{\text{goal}}\;\mathcal{U}^i,\;\mathcal{O}_k,\;r,\;R,\;\mathcal{N}^i_a,\;\mathcal{N}^i_o\rangle$ where the superscript $i$ denotes the $i^{\text{th}}$ agent, $N$ is the number of agents, $M$ is the number of obstacles, $\mathcal{X}\subseteq\mathrm{R}^n$ is the state space, $\mathcal{X}_{\text{goal}}:=\{\cup_{i=1}^N\bx_{\text{goal}}^i\in\mathcal{X}\}$ is the set consisting of non-colliding goal states,  $\mathcal{U}^i\in\mathbb{R}^m$ is the control space for the $i^{\text{th}}$ agent, $\mathcal{O}_k\subset \mathbb{R}^n$ for $k \in [1;M]_d$ is the $k^{\text{th}}$ obstacle space, $r>0$ is such that the agents must keep a safe distance from the obstacles and also among themselves i.e. $\|\bx^i-\bx^j\|\geq 2r$, where $i\neq j$, $i,j\in V$, $R>0$ is the limiting sensing radius for each agent, $\mathcal{N}^i_a=\{j \in V \mid \|\bx^i-\bx^j\| \leq R, j \neq i\}$ is the set of neighboring agents for agent $i$ and $\mathcal{N}^i_o$ is the set of neighboring obstacles for agent $i$.
 Each agent's motion is governed by the ordinary differential equation (ODE)  $\dot{\bx}^i=f(\bx^i, \bu^i)$, 
where $f:\mathcal{X}\times\mathcal{U}\rightarrow\mathcal{X}$ is a continuously differentiable function,  $\bx^i\in\mathcal{X}$ and $\bu^i \in\mathcal{U}^i$ the control input of the $i^\text{th}$ agent. 

The notion of Integral Control Barrier Function (ICBF) allows one to guarantee safety as well as input constraint satisfaction.
For agent $i$, consider the integral feedback control defined as $\dot{\bu}=\phi(\bx, \bu)$, with $\bu(0):=\bu_0\in\mathcal{U}$, where $\phi: \mathbb{R}^n\times\mathbb{R}^m \rightarrow \mathbb{R}^m$ is a given smooth function. We define the safety set $\mathcal{S}:=\{(\bx, \bu)\in\mathcal{S}_{\bx}\times\mathcal{U}\}$, incorporating both state $\mathcal{S}_\bx$ and input constraints $\mathcal{U}$. Now, consider a continuously differentiable scalar-valued function $h:\mathbb{R}^n\times\mathbb{R}^m\rightarrow\mathbb{R}$ defined such that $h(\bx, \bu)> 0$ if $(\bx, \bu)\in\mathcal{S}$, $h(\bx,\bu)=0$ if $(\bx, \bu)\in\partial\mathcal{S}$ and $h(\bx,\bu)<0$ if $(\bx, \bu)\in(\mathbb{R}^n\times\mathbb{R}^m)\setminus\mathcal{S}$. To guarantee that $(\bx,\bu)\in\mathcal{S}$, $h(\bx,\bu)$ must satisfy the forward invariance condition i.e. $\dot{h}(\bx,\bu)+\alpha(h(\bx,\bu)) \geq 0$ where $\dot{h}(\bx,\bu)$ is computed along the system trajectories and $\alpha$ is a $\mathcal{K}_\infty$ function. Formally, ICBF is defined as follows:
\begin{definition}
 \normalfont \textbf{(ICBF)} \cite{ames2020integral_ames} An Integral Control Barrier Function (ICBF) is defined as a function $h:\mathbb{R}^n\times\mathbb{R}^m\rightarrow\mathbb{R}$ that characterizes a safe set $\mathcal{S}$. Then, $h$ is said to be an ICBF, for all $(\bx, \bu)\in\mathcal{S}$ if $\bp(\bx, \bu)=0$, implies that $q(\bx, \bu) \leq 0$, where $q(\bx, \bu):  =-\left({\nabla_{\bx} h(\bx,\bu)} f(\bx, \bu)+{\nabla_{\bu} h(\bx,\bu)} \phi(\bx, \bu) +\alpha(h(\bx, \bu))\right)$ and $\bp(\bx, \bu)  :=\left({\nabla_{\bu} h(\bx,\bu)}\right)^{\mathrm{T}}$.
\label{defn:icbf}
\end{definition}
The following theorem provides a method to synthesize safe controllers via ICBF's
\begin{theorem}
    \normalfont \cite{ames2020integral_ames} If an integral feedback controller $\phi(\bx,\bu)$ and a safety set $\mathcal{S}$, defined by an ICBF $h(\bx,\bu)$, exist, then modifying the integral controller to $\dot{\bu}=\phi(\bx, \bu)+\bv^\star(\bx, \bu)$,
where $\bv^\star(\bx, \bu)$ is obtained by solving the following quadratic program (QP):
\begin{align}
&\bv^\star(\bx, \bu) = \underset{\bv \in \mathbb{R}^m}{\operatorname{argmin}}\|\bv\|^2,\;\;\text {s.t.}\;\; \bp(\bx, \bu)^{\mathrm{T}} \bv \geq q(\bx, \bu)
\end{align}
guarantees safety, i.e., forward invariance of the set $\mathcal{S}$.
\end{theorem}
\subsection{Problem statement}
Formally, in this paper, we consider the following problem
\begin{problem}
    \normalfont Given $<N,\;M,\;\mathcal{X},\;\mathcal{X}_{\text{goal}}\;\mathcal{U}^i,\;\mathcal{O}_k,\;r,\;R,$ $\;\mathcal{N}^i_a,\;\mathcal{N}^i_o>$, for each agent $i\in V$, design a decentralized controller $\bu^i(\bx^i, \bar{\bx}^i, \bar{\by}^i, \bx^i_{\text{goal}})$ where $\bar{\bx}^i$ represents the relative aggregated states of neighboring agents $\mathcal{N}^i_a$, $\bar{\by}^i$ the relative aggregated observations from neighboring obstacles $\mathcal{N}^i_o$ and $\bx^i_{\text{goal}}\in\mathcal{X}_{\text{goal}}$ such that the following holds true: 
    \begin{itemize}
    \item Input constraint satisfaction: Each agent's control policy satisfies the input constraints i.e. $\|\bu^i\|\leq u_{\text{max}}$ where $i\in V$ and $u_\text{max}>0$.
       \item  Guaranteed safety: No collisions with obstacles (i.e., $\|\bar{\by}^i(t)\|>R$ as well as with other agents, meaning $\|\bx^i(t)-\bx^j(t)\|>2r$ for $t \geq 0$, $r>0$ and $j \neq i$.
        \item Minimizing the number of deadlocks i.e., the phenomenon when multiple agents stop before reaching their goal.
 \end{itemize}

 
 \label{problem:main}
\end{problem}
For the sake of brevity, we drop stating the arguments of $\bu^i$ unless required. We are now in a position to discuss our proposed approach in the following section.$ $
\section{\ours: Proposed Framework\label{sec:proposed_framework}}
This section proposes our approach termed \ours that satisfies the objectives presented in Section \ref{sec:prelim_and_problem}. Our approach is primarily divided into two steps. The first step involves the joint learning of the Neural Integral Control Barrier Function (NICBF) certificates and a decentralized safe control policy that enables the $N$ agents to reach their respective goals. Since the control policies are parameterized by neural networks, they do not provide complete safety guarantees for the agents. Furthermore, as $N$ increases, the likelihood of agents experiencing deadlock also increases. To address these two fundamental limitations, the second step involves developing theoretical frameworks that is augmented to the NN policy (in the first step) for avoiding collisions and minimizing deadlocks. To provide complete guarantees on the collision avoidance problem, we shift to a decentralized optimization-based control policy termed MPC-ICBF whenever we predict a collision in future time steps with its neighbors using the learned NN policy. Next, we analyze the deadlock problem by writing the multi-agent control problem as a QP. Then, we write the KKT conditions that provide sufficient conditions for the optimal control problem. Consequently, we draw an analogy between this deadlock problem and the problem of minimizing non-convex loss functions. Finally, we propose a method to modify the control input to ensure that the deadlocks are minimized.

\subsection{Learning framework\label{subsec:learning_framework}}
Given the constants from Problem \ref{problem:main}, the task for an agent $i\in V$ is to synthesize an ICBF $h^i_{\theta_h}(\bx,\bu)$ and a decentralized feedback controller {$\pi^i_{\theta_c}\left(\bx^i, \bar{\bx}^i, \bar{\by}^i, \bx^i_{\text{goal}}\right)$} (where $\theta_h$ and $\theta_c$ are the parameters of the NN) that would take the agents towards their respective goal positions $\bx^i_{\text{goal}}\in\mathcal{X}_{\text{goal}}$. { Motivated by the fact that the neural network architecture for $h_{\theta_h}^i(\bx,\bu)$ must adapt to changes in the size of $\bar{\by}^i$ and must be permutation invariant with respect to the neighboring agents $\mathcal{N}^i_a$, we use an architecture similar to that in \cite{qi2017pointnet}. 
Consider a time-varying observation matrix $\bar{\by}^i(t) \in \mathbb{R}^{n \times\left|\mathcal{N}^i_a\right|}$. Let $W \in \mathbb{R}^{p \times n}$ be the weight matrix and $\sigma(\cdot)$ the element-wise $\operatorname{ReLU}$ activation function. The mapping $\rho: \mathbb{R}^{n \times\left|\mathcal{N}^i_a\right|} \mapsto \mathbb{R}^p$ is given by
$
\rho\left(\bar{\by}^i\right)=\operatorname{RowMax}\left(\sigma\left(W \bar{\by}^i\right)\right),
$
where $\rho$ maps a matrix $\bar{\by}^i$ with a dynamic column dimension and permutation to a fixed-length feature vector $\rho\left(\bar{\by}^i\right) \in \mathbb{R}^p$ and  $\operatorname{RowMax}(\cdot)$ as the row-wise max pooling operation. The dimension of $\rho\left(\bar{\by}^i\right)$ remains constant even if $\left|\mathcal{N}^i_a\right|$ changes over time. Similarly, the ICBF is parameterized, with the output being a scalar instead of a vector.} 
The main objective is to ensure that the learned ICBF $h^i_\theta(\bx,\bu)$ satisfies the ICBF based conditions given in Definition \ref{defn:icbf}. 
Towards this goal, we consider minimizing the following loss function $\mathcal{L}=\sum_{j=1}^N \mathcal{L}^j_h(\theta)$ where $ \mathcal{L}^j_h(\theta)$ for agent $j\in V$ $\left(\theta=\theta_c\cup\theta_h\right)$:
\begin{align}
&\mathcal{L}^j_h(\theta) = \sum_{i=1}^{N_h} \left(-p_\theta\left(\bx^{j,\text{S}}_i, \bu^{j,\text{S}}_i\right)^\mathrm{T} \pi^j_{\theta_c} + q_{\theta}\left(\bx^{j,\text{S}}_i, \bu^{j,\text{S}}_i\right) \right)_++ \nonumber\\
& \sum_{i=1}^{N_s}\left(-h^j_{\theta_h}\left(\bx^{j,\text{S}}_i, \bu^{j,\text{S}}_i\right)\right)_+ + \sum_{i=1}^{N_h - N_s}\left(h^j_{\theta_h}\left(\bx^{j,\text{US}}_i, \bu^{j,\text{US}}_i\right)\right)_+
\label{eqn:loss_h}
\end{align}
where $ x_+ = \max\{0, x\} $, $ N_h $ and $N_s(< N_h)$ denote the number of total state-input pairs and safe state-input pairs respectively, $\bx^{j,\text{S}}_i \in {\mathcal{S}^j_\bx}$, $\bx^{j,\text{US}}_i \in \mathcal{X} \setminus \mathcal{S}^j_\bx$, $\bu^{j,\text{S}}_i \in \mathcal{U}$, $\bu^{j,\text{US}}_i \notin \mathcal{U}$,
and $ p_\theta\left(\bx^j, \pi^j_{\theta_c}\right) $ and $ q_\theta\left(\bx^j, \pi^j_{\theta_c}\right) $ are defined as:
\begin{align}
&p_\theta\left(\bx^j, \pi^j_{\theta_c}\right) := \left(\nabla_{\pi^j_\theta} h^j_{\theta_h}\left(\bx^j, \pi^j_\theta\right)\right)^{\mathrm{T}},\nonumber\\
&q_{\theta}\left(\bx^j, \pi^j_{\theta_c}\right) := -\left(\nabla_{\bx^j} h^j_{\theta_h}\left(\bx^j, \pi^j_{\theta_c}\right) f\left(\bx^j, \pi^j_{\theta_c}\right)\right.\nonumber\\
&\left.\quad\;+ \nabla_{\pi^j_\theta} h^j_{\theta_h}\left(\bx^j, \pi^j_{\theta_c}\right) \phi\left(\bx^j, \pi^j_{\theta_c}\right) + \alpha\left(h^j_{\theta_h}\left(\bx^j, \pi^j_{\theta_c}\right)\right)\right)\nonumber
\end{align}
where $\phi$ is an integral controller given by $\phi=\dot{k}(\bx)=\nabla_\bx k(\bx)f(\bx,\bu)$ and $k(\bx)$ is a nominal controller (for instance LQR or PID controller).

\textbf{Data collection:} Note that we do not use a fixed set of state-input pairs to train the decentralized ICBF $h^j_{\theta_h}$ and feedback controllers $\pi^j_\theta$. Instead, we adopt an online training strategy, collecting data by running the current system. These pairs are temporarily stored to calculate loss and update the controllers via gradient descent. The updated controllers then generate new training data. This iterative process continues until the loss converges.
\subsection{Deadlock analysis, deadlock resolution, and collision avoidance under limited actuation\label{subsec:collision_avoidance_and_deadlock_analysis}}
In this section, we perform a deadlock analysis assuming that the governing dynamics is control-affine, i.e., $\dot{\bx}=f(\bx)+g(\bx)\bu$ with state constraints only. However, a similar analysis holds for non-control affine systems with input constraints as well. Note that each agent $i$ computes its control input by solving the following QP (dropped the superscript $i$ for brevity):
\begin{align}
    \bu^\star=
    \underset{\bu\in\mathbb{R}^m}{\operatorname{argmin}}\; & \|\bu - k(\bx)\|_2^2 \quad\quad\text{s.t. } \; \bar{A} \bu \leq \bar{\boldsymbol{q}}
    \label{eqn:qp_analysis}
\end{align}
where $\bar{A}$ and $\overline{\boldsymbol{q}}$ are given by
\begin{align}
    &\bar{A}=[L_gh^1(\bx),\dots,L_gh^{i-1}(\bx),L_gh^{i+1}(\bx),\dots L_gh^N(\bx)]\nonumber\\
    &\overline{\boldsymbol{q}}=[L_fh^1(\bx)+\alpha(h^1(\bx)),\dots,L_fh^{i-1}(\bx)+\alpha(h^{i-1}(\bx)),\nonumber\\
    &\quad\quad L_fh^{i+1}(\bx)+\alpha(h^{i+1}(\bx)),\dots ,L_fh^N(\bx)+\alpha (h^N(\bx))]\nonumber
\end{align}
The superscript in $h^i(\bx)$  denotes the CBF for the $i^\text{th}$ agent. 
The notation $\overline{\boldsymbol{p}}_k$ represents the $k^{\text{th}}$ row of $\bar{A}$, and $\tilde{q}_k$ the $k^\text{th}$ component of $\overline{\boldsymbol{q}}$. The Lagrangian associated with \eqref{eqn:qp_analysis} is
$L(\bu, \boldsymbol{\lambda}) = \|\bu - k(\bx)\|_2^2 + \sum_{k=1}^{N} \lambda_k(\tilde{\boldsymbol{p}}_k^\mathrm{T} \bu - \tilde{q}_k)$.
At optimality, the pair $(\bu^\star, \boldsymbol{\lambda}^\star)$ satisfies the Karush-Kuhn-Tucker (KKT) conditions as follows:
\begin{enumerate}
    \item \textbf{Stationarity}: The gradient of the Lagrangian $L$ with respect to $\bu$ at $(\bu^\star, \boldsymbol{\lambda}^\star)$ is zero, leading to
$
\bu^\star = k(\bx) - \frac{1}{2} \sum_{k=1}^{N} \lambda_k^\star \tilde{\boldsymbol{p}}_k^\mathrm{T}
$.
\item \textbf{Primal Feasibility}: This requires that 
$\overline{\boldsymbol{p}}_k^\mathrm{T} \bu^\star \leq \tilde{q}_k $ for all $ k \in V$. 
\item \textbf{Dual Feasibility}: 
$
\lambda_k^\star \geq 0 \, 
$ for all $k \in V$.
\item \textbf{Complementary Slackness}: This ensures that for all constraints,  
$
\lambda_k^\star (\tilde{\boldsymbol{p}}_k^\mathrm{T} \bu^\star - \bar{q}_k) = 0 $ for all $ k \in V
$.

Mathematically, 
$
\bu^\star = k(\bx) - \frac{1}{2} \sum_{k \in \mathcal{A}(\bu^\star)} \lambda_k^\star \tilde{\boldsymbol{p}}_k^\mathrm{T}
$ where $\mathcal{A}(\bu^\star)  = \{k \in V \mid \tilde{\boldsymbol{p}}_k^\mathrm{T} \bu^\star = \tilde{q}_k\}$ is the set of active constraints .
\end{enumerate}
The system is in a deadlock when $\bu^\star$ is zero i.e. $k(\bx) = \frac{1}{2} \sum_{k \in \mathcal{A}(\bu^\star)} \lambda_k^\star \tilde{\boldsymbol{p}}_k^\mathrm{T}$. This deadlock phenomenon of agents getting stuck in local minima (due to non-convexity of $k(\bx)$ wrt to $\bx$ in \eqref{eqn:qp_analysis}) is similar to the problem of minimizing non-convex functions in ML using well-established gradient methods.
In the following, inspired by gradient based methods in ML that minimize a non-convex loss function, we propose a deadlock minimization method to minimize the deadlock present in large scale multi-agent control.

\textbf{Deadlock resolution:}\label{deadlock_mechanism} Note that the cost function \eqref{eqn:qp_analysis} is quadratic in $\bu$, however as $\bu$ in \eqref{eqn:qp_analysis} is a function of the state $\bx$, the cost function can be non-convex with respect to $\bx$. The deadlock occurs when the agents gets stuck in the local minima of this cost function i.e. when the cost becomes $\|k(\bx)\|_2^2$ (note that $\bx$ is fixed when solving the QP \eqref{eqn:qp_analysis}) instead of the global minima, before it reaches the goal. To avoid these local minima, we leverage tools from gradient-based methods in ML to escape the local minima. Particularly, in this paper, we leverage the momentum based gradient method to modify $\bu^\star_t$ from \eqref{eqn:qp_analysis} as follows (subscript $t$ denotes the time instant at which \eqref{eqn:qp_analysis} is solved):
\begin{align}
   & \bu^\star_t=\bu^\star_{t-1}-\boldsymbol{w}_t+\Delta^\star,\quad \boldsymbol{w}_t=\gamma \boldsymbol{w}_{t-1}+\eta \nabla_\bx c(\bx;\theta)
    \label{eqn:deadlock_resolve}
\end{align}
where $\eta$ is the learning rate, $\gamma\in(0,1)$, $c(\bx;\theta)=\|\bu(\bx;\theta_c) - k(\bx)\|_2^2$. As we know, the sequence of the states and the inputs of our neighboring agents at some previous time steps, it is possible to approximately compute the values of $h_{t-1}(\bx;\theta)$ for the agents. Please note that the input computed using the update scheme would be deadlock-free, however, there are no guarantees whether the agents are collision-free. Towards this goal, we choose the value of $\Delta^\star$ in \eqref{eqn:deadlock_resolve} as follows:
\begin{align}
\Delta^\star=&\underset{\Delta\in\mathbb{R}^m}{\operatorname{argmin}}\;\;\|\Delta\|^2\quad\quad\text{s.t. } \bar{A}\bu^\star_t\leq \Bar{\boldsymbol{q}}
\label{eqn:qp_deadlock_free}
\end{align}
where $\bu^\star_t$ is from \eqref{eqn:deadlock_resolve}. Note that \eqref{eqn:qp_deadlock_free} is a QP as the optimization problem is quadratic with respect to the variable $\Delta$. Updating the control input via \eqref{eqn:deadlock_resolve} is only triggered when the magnitude of the control input $\bu^{\star}_t$ of at least one of the agents is below a threshold i.e. $\|\bu^{\star}_t\|\leq \epsilon$ for some small $\epsilon>0$. To escape the local minima, state-of-the-art gradient-based update schemes can be used to update the control input at time $t$ in \eqref{eqn:deadlock_resolve}.

\textbf{Safety guarantee}: One of the trivial methods to guarantee safety is to embed an optimization based framework such as MPC whenever a collision is anticipated by forward propagating the agent’s dynamics for a few time steps via the NN based controller. The MPC problem when a collision is anticipated (via the NN control policy) is given by
\begin{subequations}
    \begin{align}
    \bu_i^\star=\;&\underset{}{\text{argmin}}\;c_{\text{MPC}}(\bx,\bu)\\
    \text{s.t.}\quad& h^j(\bx)+\alpha(h^j(\bx))\geq 0\quad\forall\quad j\in\mathcal{N}^i_a\\
    & h^j(\bu)+\alpha(h^j(\bu))\geq 0\quad\forall\quad j\in\mathcal{N}^i_a
\end{align}
    \label{eqn:mpc_trivial}
\end{subequations}
where $c_{\text{MPC}}(\bx,\bu)>0$ is the cost function, usually taken to be a quadratic function, and $h^j(\bx)$ and $h^j(\bu)$ ($j\in\mathcal{N}^i_a$) encodes the state and the input constraints respectively. With a slight abuse in notation, we interchangeably change the inputs for $h^j$ to $\bx$ (only state constraints) or $\bu$ (only input constraints) or both $(\bx,\bu)$. However, the optimization problem is computationally expensive especially when the number of agents are large (as the number of constraints are $N-1$), thereby resulting in poor scalability. To address this limitation, we transform the $2(N-1)$ constraints into a single constraint, thereby making the MPC problem devoid of the number of the agents. Particularly, we consider the combined ICBF $h^i_c(\bx,\bu)$ for an agent $i$ as follows:
\begin{align}
    h^i_c(\bx,\bu)=-\text{log}\left(\sum_{j\in\mathcal{N}^i_a}e^{-\beta h^j(\bx)} + e^{-\beta h^j(\bu)}\right)
    \label{eqn:icbf}
\end{align}
where $\beta>0$. Consequently, the MPC-ICBF with a single constraint becomes
\begin{align}
\bu_i^\star=\underset{\bu\in\mathcal{U}^i}{\text{argmin}}\;c_\text{MPC}(\bx,\bu)\;\;
    \text{s.t.}\;\; \dot{h}^i_c(\bx,\bu)+\alpha(h^i_c(\bx,\bu))\geq 0
    \label{eqn:mpc_proposed_via_icbf}
\end{align}
Consider the following three constraint sets for agent $i$, i.e.,
\begin{subequations}
    \begin{align}
    & \mathcal{S}_\bx^i=\left\{\bx:\;\cap_{j\in\mathcal{N}^i_a}h^j(\bx)+\alpha(h^j(\bx))\geq 0\right\}\\
        & \mathcal{S}_\bu^i=\left\{\bu:\;\cap_{j\in\mathcal{N}^i_a}h^j(\bu)+\alpha(h^j(\bu))\geq 0\right\}\\
    &\mathcal{S}^i_c=\left\{(\bx,\bu):\;h^i_c(\bx,\bu)+\alpha(h^i_c(\bx,\bu))\geq 0\right\}
\end{align}
\end{subequations}
It can be easily shown that the set $\mathcal{S}_{\bx}^i\cup \mathcal{S}_{\bu}^i\subset\mathcal{S}^i_c$. Furthermore, $  \underset{\beta\rightarrow\infty}{\lim}\;\mathcal{S}^i_c=\mathcal{S}_{\bx}^i\cup \mathcal{S}_{\bu}^i$. For the sake of brevity, we do not present the proof in this paper.

\textbf{Input constraint satisfaction:} Based on the definition of ICBF, the combined ICBF \eqref{eqn:icbf} encodes both the state as well as the input constraints. Consequently, solving the proposed MPC-ICBF \eqref{eqn:mpc_proposed_via_icbf} ensures that the input constraints are satisfied.
\section{Evaluation and Discussion\label{sec:results}}

In this section, we compare our proposed approach with two classes of methods. Through the experiments, we aim to answer the following questions $(i)$ can \ours guarantee 100\% collision avoidance as well as scale well with the number of agents (e.g. $>$1000 agents)? $(ii)$ can \ours reduce the number of deadlocks among agents compared to baseline methods while synthesizing safe trajectories and respecting input constraints? $(iii)$ does  \ours take care of practical considerations such as input constraint satisfaction while maintaining the safety guarantee?, $(iv)$ whether \ours provides significant computational benefits versus the approach that trivially combines the NN based controller and MPC with multiple ICBF constraints \eqref{eqn:mpc_trivial} and finally $(v)$ whether \ours is generalizable. Particularly, does \ours perform well if they are trained on fewer agents and tested for a larger number of agents?





\begin{figure}[h!]


    \begin{minipage}{0.24\textwidth}
        \centering
        \begin{tikzpicture}[scale=0.35]

\definecolor{agentcolor}{RGB}{102,153,204}
\definecolor{targetcolor}{RGB}{102,204,153}

\draw[line width=0.9mm] (0,0) rectangle (10,10);

\node[circle,fill=agentcolor,inner sep=0pt,minimum size=7pt] at (2.5,8.5) {7};
\node[circle,fill=targetcolor,inner sep=0pt,minimum size=7pt] at (2,7) {5};
\node[circle,fill=agentcolor,inner sep=0pt,minimum size=7pt] at (3,6.5) {5};
\node[circle,fill=agentcolor,inner sep=0pt,minimum size=7pt] at (8.5,7) {4};
\node[circle,fill=targetcolor,inner sep=0pt,minimum size=7pt] at (1.5,5) {4};
\node[circle,fill=targetcolor,inner sep=0pt,minimum size=7pt] at (3.5,4) {2};
\node[circle,fill=agentcolor,inner sep=0pt,minimum size=7pt] at (4.5,4) {1};
\node[circle,fill=targetcolor,inner sep=0pt,minimum size=7pt] at (5.5,5) {6};
\node[circle,fill=targetcolor,inner sep=0pt,minimum size=7pt] at (7.5,5) {3};
\node[circle,fill=agentcolor,inner sep=0pt,minimum size=7pt] at (8,4.5) {2};
\node[circle,fill=agentcolor,inner sep=0pt,minimum size=7pt] at (2,2) {6};
\node[circle,fill=targetcolor,inner sep=0pt,minimum size=7pt] at (5,2.5) {1};
\node[circle,fill=agentcolor,inner sep=0pt,minimum size=7pt] at (6,2) {3};
\node[circle,fill=targetcolor,inner sep=0pt,minimum size=7pt] at (8.5,2) {7};

\node[circle,fill=agentcolor,inner sep=0pt,minimum size=7pt] at (7,9) {};
\node[right, text width=3cm] at (7.0,9) {Agent};
\node[circle,fill=targetcolor,inner sep=0pt,minimum size=7pt] at (7,8.3) {};
\node[right] at (7.0,8.3) {Target};

\end{tikzpicture}
        \subcaption{\small Empty}
        \label{fig:d}
    \end{minipage}
    \hfill
    \begin{minipage}{0.24\textwidth}
        \centering
        \begin{tikzpicture}[scale=0.4]
\draw[lightgray!20] (0,0) grid (8,8);

\draw[line width=3pt,
cap=round,
rounded corners=1pt,
draw=black!75] (-0.5,-0.5) -| (8.5,8.5)
(7.5,8.5)-|(-0.5,0.5) -- (0.5,0.5) (-0.5,2.5)-|(0.5,4.5)
(0.5,5.5)|-(4.5,7.5)|-(3.5,3.5)|-(2.5,2.5)|-(3.5,1.5)
(8.5,0.5)--(6.5,0.5)
(5.5,0.5)-|(3.5,-0.5)
(3.5,0.5)-|(1.5,3.5)-|(2.5,5.5)
(2.5,4.5)-|(3.5,6.5)-|(1.5,4.5)
(7.5,8.5)|-(5.5,6.5)--(5.5,7.5)
(6.5,7.5)--(6.5,8.5)
(8.5,3.5)-|(6.5,4.5)-|(7.5,5.5)
(6.5,5.5)-|(5.5,2.5)--(4.5,2.5)
(5.5,2.5)-|(7.5,1.5)--(4.5,1.5)
(5.5,4.5)--(4.5,4.5)
(1.5,1.5)--(0.5,1.5)
(1.5,7.5)--(1.5,8.5);
\definecolor{agentcolor}{RGB}{102,153,204}
\definecolor{targetcolor}{RGB}{102,204,153}
\node[circle,fill=agentcolor,inner sep=0pt,minimum size=7pt] at (5.0,6.5) {0};
\node[circle,fill=agentcolor,inner sep=0pt,minimum size=7pt] at (8.0,7) {4};
\node[circle,fill=targetcolor,inner sep=0pt,minimum size=7pt] at (1.0,5) {4};
\node[circle,fill=targetcolor,inner sep=0pt,minimum size=7pt] at (3.5,4) {2};

\node[circle,fill=agentcolor,inner sep=0pt,minimum size=7pt] at (4.0,4.9) {1};
\node[circle,fill=targetcolor,inner sep=0pt,minimum size=7pt] at (5.0,5) {6};
\node[circle,fill=targetcolor,inner sep=0pt,minimum size=7pt] at (7.0,5) {3};
\node[circle,fill=agentcolor,inner sep=0pt,minimum size=7pt] at (8,4.5) {2};
\node[circle,fill=agentcolor,inner sep=0pt,minimum size=7pt] at (1.0,2) {6};
\node[circle,fill=targetcolor,inner sep=0pt,minimum size=7pt] at (3.0,3) {1};
\node[circle,fill=agentcolor,inner sep=0pt,minimum size=7pt] at (6,2) {3};
\node[circle,fill=targetcolor,inner sep=0pt,minimum size=7pt] at (2,7) {5};
\node[circle,fill=agentcolor,inner sep=0pt,minimum size=7pt] at (3,6.0) {5};
\end{tikzpicture}
        \subcaption{\small Maze}
        \label{fig:d1}
    \end{minipage}
\caption{\small Representative testing environments used for experiments. (a) Empty environment. (b) Complex maze environment.
}
    \label{fig:envs}
    \vspace{-10pt}
\end{figure}
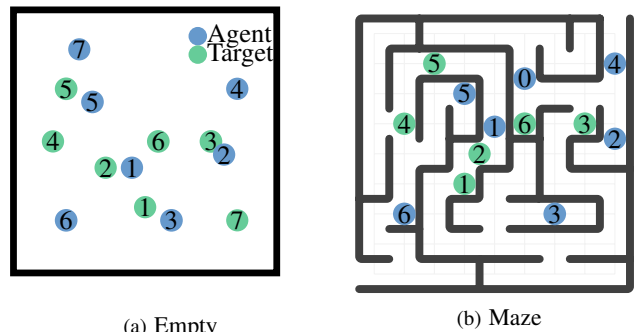
\subsection{Experimental Setup}
\label{subsec: experimental_setup}
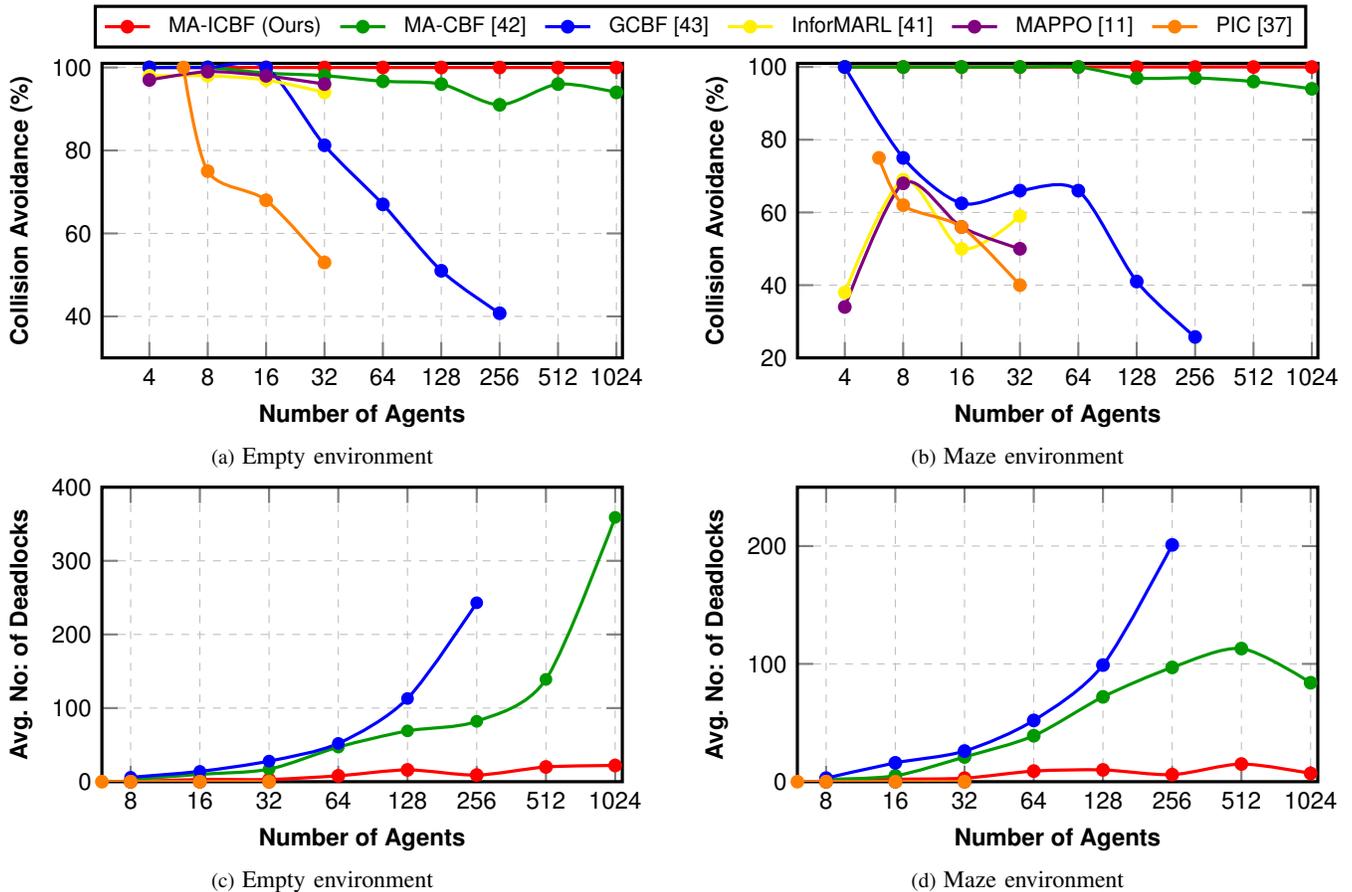
\begin{figure*}[hthp]
    \centering
    \begin{subfigure}[b]{0.48\textwidth}
        \centering
    \captionsetup{skip=-5pt}





\begin{tikzpicture}
\begin{axis}[
scale=\scaletikz,
    title style={font=\large\sansmath\sffamily, yshift=1ex},
    xlabel={\agentempty},
    ylabel={\textbf{Collision Avoidance (\%)}},
    width=\widthtikz, 
    height=\heighttikz, 
    xtick={ 4, 8, 16, 32, 64, 128, 256, 512, 1024},
    xticklabels={ 4, 8, 16, 32, 64, 128, 256, 512, 1024},
    legend pos=south west,
    legend cell align={left},
    legend style={font=\small\sansmath\sffamily},
    xmajorgrids=true,
    ymajorgrids=true,
    grid style=dashed,
    font=\small\sansmath\sffamily,
    tick style={major tick length=6pt, minor tick length=3pt, thick},
    xmode=log,
    log basis x={2},
    every axis plot/.append style={thick},
    scaled ticks=false,
    xmin=0, xmax=1100,
    ymin=30, ymax=101,
    cycle list name=color list,
    line width=1.2pt,
        legend style={
        at={(1.15,1.05)},
        anchor=south,
        legend columns=6,
        font=\small\sansmath\sffamily,
        column sep=1ex
    },
    legend cell align={left},
]

\addplot[
    color=red,
                    mark=*,
    mark options={solid, fill=red},
    line width=1.2pt,
    smooth
    ]
    coordinates {
      (4,100)  (8,100) (16,100) (32,100) (64,100) (128,100) (256,100) (512,100) (1024,100)
    };
\addlegendentry{\footnotesize \ours (Ours)}

\addplot[
    color=green!60!black,
                    mark=*,
    mark options={solid, fill=green!60!black},
    line width=1.2pt,
    smooth
    ]
    coordinates {
       (4,100)  (8,100) (16,98.66) (32,98.03) (64, 96.68) (128,96) (256,91) (512,96) (1024,94)
    };
\addlegendentry{\footnotesize MA-CBF\cite{qin2021learning_macbf}}

\addplot[
    color=blue,
    mark=*,
    mark options={solid, fill=blue},
    line width=1.2pt,
    smooth
    ]
    coordinates {
     (4,100) (8,100) (16,100) (32,81.25) (64,67) (128, 51) (256, 40.75)
    };
\addlegendentry{\footnotesize GCBF\cite{zhang2023neural_gcbf}}

\addplot[
    color=yellow,
                    mark=*,
    mark options={solid, fill=yellow},
    line width=1.2pt,
    smooth
    ]
    coordinates {
     (4,98) (8,98) (16,97) (32,94) 
    };
\addlegendentry{\footnotesize InforMARL\cite{nayak22informarl}}

\addplot[
    color=violet,
                    mark=*,
    mark options={solid, fill=violet},
    line width=1.2pt,
    smooth
    ]
    coordinates {
      (4,97) (8,99) (16,98) (32,96)
    };
\addlegendentry{\footnotesize MAPPO \cite{yu2022surprising_ma_lit_1}}

\addplot[
    color=orange,
                    mark=*,
    mark options={solid, fill=orange},
    line width=1.2pt,
    smooth
    ]
    coordinates {
     (6,100) (8,75) (16,68) (32,53) 
    };
\addlegendentry{\footnotesize PIC \cite{liu2020pic}} 

\end{axis}
\end{tikzpicture}
        \caption{\small Empty environment}        \label{fig:collision_empty}
    \end{subfigure}
    \hfill
    \begin{subfigure}[b]{0.48\textwidth}
        \centering
    \captionsetup{skip=-5pt}
            \vspace{-0.5cm}





\begin{tikzpicture}
\begin{axis}[
scale=\scaletikz,
    title style={font=\large\sansmath\sffamily, yshift=1ex},
    xlabel={\agentmaze},
    ylabel={\textbf{Collision Avoidance (\%)}},
    width=\widthtikz, 
    height=\heighttikz, 
    xtick={4, 8, 16, 32, 64, 128, 256, 512, 1024},
    xticklabels={4, 8, 16, 32, 64, 128, 256, 512, 1024},
    xmajorgrids=true,
    ymajorgrids=true,
    grid style=dashed,
    font=\small\sansmath\sffamily,
    tick style={major tick length=6pt, minor tick length=3pt, thick},
    xmode=log,
    log basis x={2},
    every axis plot/.append style={thick},
    scaled ticks=false,
    xmin=0, xmax=1100,
    ymin=20, ymax=101,
    cycle list name=color list,
    line width=1.2pt,
]

\addplot[
    color=red,
                    mark=*,
    mark options={solid, fill=red},
    line width=1.2pt,
    smooth
    ]
    coordinates {
      (4,100)  (8,100) (16,100) (32,100) (64,100) (128,100) (256,100) (512,100) (1024,100)
    };

\addplot[
    color=green!60!black,
                mark=*,
    mark options={solid, fill=green!60!black},
    line width=1.2pt,
    smooth
    ]
    coordinates {
       (4,100)  (8,100) (16,100) (32,100) (64,100) (128,97) (256,97) (512,96) (1024,94)
    };

\addplot[
    color=blue,
                    mark=*,
    mark options={solid, fill=blue},
    line width=1.2pt,
    smooth
    ]
    coordinates {
     (4,100) (8,75) (16,62.5) (32,66) (64,66) (128, 41) (256, 25.75)
    };
\addplot[
    color=yellow,
                    mark=*,
    mark options={solid, fill=yellow},
    line width=1.2pt,
    smooth
    ]
    coordinates {
     (4,38) (8,69) (16,50) (32,59) 
    };

\addplot[
    color=violet,
                    mark=*,
    mark options={solid, fill=violet},
    line width=1.2pt,
    smooth
    ]
    coordinates {
     (4,34) (8,68) (16,56) (32,50) 
    };
\addplot[
    color=orange,
                    mark=*,
    mark options={solid, fill=orange},
    line width=1.2pt,
    smooth
    ]
    coordinates {
     (6,75) (8,62) (16,56) (32,40) 
    };

\end{axis}
\end{tikzpicture}
        \caption{\small Maze environment}
\label{fig:collision_avoidance}
    \end{subfigure}
    \centering
    \begin{subfigure}[b]{0.48\textwidth}
        \centering
    \captionsetup{skip=-5pt}





\begin{tikzpicture}
\begin{axis}[
scale=\scaletikz,
    title style={font=\large\sansmath\sffamily, yshift=1ex},
    xlabel={\agentempty},
    ylabel={\textbf{Avg. No: of Deadlocks}},
    width=\widthtikz, 
    height=\heighttikz, 
    xtick={8, 16, 32, 64, 128, 256, 512, 1024},
    xticklabels={8, 16, 32, 64, 128, 256, 512, 1024},
    legend pos=north west,
    legend cell align={left},
    legend style={font=\small\sansmath\sffamily},
    xmajorgrids=true,
    ymajorgrids=true,
    grid style=dashed,
    font=\small\sansmath\sffamily,
    tick style={major tick length=6pt, minor tick length=3pt, thick},
    xmode=log,
    log basis x={2},
    every axis plot/.append style={thick},
    scaled ticks=false,
    xmin=6, xmax=1100,
    ymin=0, ymax=400,
    cycle list name=color list,
    line width=1.2pt,
        legend style={
        at={(1.3,1.1)},
        anchor=south,
        legend columns=6,
        font=\small\sansmath\sffamily,
        column sep=1ex
    },
    legend cell align={left},
]

\addplot[
    color=red,
                    mark=*,
    mark options={solid, fill=red},
    line width=1.2pt,
    smooth
    ]
    coordinates {
        (8, 0) (16, 3) (32,3) (64,8) (128,16) (256,9) (512,20) (1024,22)
    };

\addplot[
    color=green!60!black,
                mark=*,
    mark options={solid, fill=green!60!black},
    mark size=1.8,
    line width=1.2pt,
    smooth
    ]
    coordinates {
    (8, 3) (16, 10) (32,17) (64,47) (128,69) (256, 82) (512,139) (1024,359)
    };

\addplot[
    color=blue,
            mark=*,
    mark options={solid, fill=blue},
    mark size=1.8,
    line width=1.2pt,
    smooth
    ]
    coordinates {
        (8, 6) (16, 14) (32,28) (64,52) (128, 113) (256, 243)
    };

\addplot[
    color=yellow,
                    mark=*,
    mark options={solid, fill=yellow},
    line width=1.2pt,
    smooth
    ]
    coordinates {
        (4,0) (8,0) (16,0) (32,0) 
    };

\addplot[
    color=violet,
                    mark=*,
    mark options={solid, fill=violet},
    line width=1.2pt,
    smooth
    ]
    coordinates {
        (4,0) (8,0) (16,0) (32,0) 
    };

\addplot[
    color=orange,
                    mark=*,
    mark options={solid, fill=orange},
    line width=1.2pt,
    smooth
    ]
    coordinates {
        (6,0) (8,0) (16,0) (32,0) 
    };

\end{axis}
\end{tikzpicture}
        \caption{\small Empty environment}
    \label{fig:deadlock_empty}
    \end{subfigure}
    \hfill
    \begin{subfigure}[b]{0.48\textwidth}
        \centering
    \captionsetup{skip=-5pt}





\begin{tikzpicture}
\begin{axis}[
scale=\scaletikz,
    title style={font=\large\sansmath\sffamily, yshift=1ex},
    xlabel={\agentmaze},
    ylabel={\textbf{Avg. No: of Deadlocks}},
    width=\widthtikz, 
    height=\heighttikz, 
    xtick={8, 16, 32, 64, 128, 256, 512, 1024},
    xticklabels={8, 16, 32, 64, 128, 256, 512, 1024},
    legend pos=north west,
    legend cell align={left},
    legend style={font=\small\sansmath\sffamily},
    xmajorgrids=true,
    ymajorgrids=true,
    grid style=dashed,
    font=\small\sansmath\sffamily,
    tick style={major tick length=6pt, minor tick length=3pt, thick},
    xmode=log,
    log basis x={2},
    every axis plot/.append style={thick},
    scaled ticks=false,
    xmin=6, xmax=1100,
    ymin=0, ymax=250,
    cycle list name=color list,
    line width=1.2pt,
        legend style={
        at={(0.42,1.1)},
        anchor=south,
        legend columns=3,
        font=\small\sansmath\sffamily,
        column sep=1ex
    },
    legend cell align={left},
]

\addplot[
    color=red,
                    mark=*,
    mark options={solid, fill=red},
    line width=1.2pt,
    smooth
    ]
    coordinates {
        (8, 0) (16, 2) (32,3) (64,9) (128,10) (256,6) (512,15) (1024,7)
    };

\addplot[
    color=green!60!black,
                    mark=*,
    mark options={solid, fill=green!60!black},
    line width=1.2pt,
    smooth
    ]
    coordinates {
    (8, 2) (16, 5) (32,21) (64,39) (128,72) (256, 97) (512,113) (1024,84)
    };

\addplot[
    color=blue,
                    mark=*,
    mark options={solid, fill=blue},
    line width=1.2pt,
    smooth
    ]
    coordinates {
        (8, 3) (16, 16) (32,26) (64,52) (128, 99) (256, 201)
    };

\addplot[
    color=yellow,
                    mark=*,
    mark options={solid, fill=yellow},
    line width=1.2pt,
    smooth
    ]
    coordinates {
        (4,0) (8,0) (16,0) (32,0) 
    };

\addplot[
    color=violet,
                    mark=*,
    mark options={solid, fill=violet},
    line width=1.2pt,
    smooth
    ]
    coordinates {
        (4,0) (8,0) (16,0) (32,0) 
    };

\addplot[
    color=orange,
                    mark=*,
    mark options={solid, fill=orange},
    line width=1.2pt,
    smooth
    ]
    coordinates {
        (6,0) (8,0) (16,0) (32,0) 
    };
\end{axis}
\end{tikzpicture}
        \caption{\small Maze environment}
    \label{fig:deadlock_maze}
    \end{subfigure}
    \caption{\small (a)-(b): Collision avoidance percentage versus number of agents. (c)-(d): Average number of deadlocks.}
    \label{fig:deadlock_envs}
    \vspace{-10pt}
\end{figure*}
For the experiments, we consider an empty environment and a complex maze environment which is a $10m \times 10m$ grid as shown in Fig.~\ref{fig:envs}. We consider a quadrotor system with dynamics similar to that in \cite{qin2021learning_macbf}. The training is performed with $70000$ timesteps with a learning rate of $0.001$. For each experiments 5 trials are conducted with each trial being randomized  with respect to start and goal position in both of the environments. All experiments have been conducted on an AMD Ryzen machine @$2.90$GHz with $8$ cores and $16.0$ GB of RAM. The maximum number of agents tested in our experiments is $1024$. First, we consider multi-agent reinforcement learning (MARL) algorithms~\cite{liu2020pic,yu2022surprising_mappo,nayak22informarl}. The main limitation of these MARL algorithms is that they do not exploit the nonlinear dynamics of the system and henceforth require a large number of state-input pairs to train their algorithm.
Due to computational intractability, we test the MARL algorithms only up to $32$ agents similar to \cite{qin2021learning_macbf,zhang2023neural_gcbf} and up to $256$ agents for method \cite{zhang2023neural_gcbf}. Second, we consider neural network-based (NN-based) approaches that leverage CBF-based safety certificates and exploit the nonlinear dynamics to design safe control inputs \cite{qin2021learning_macbf,zhang2023neural_gcbf}.
We choose $\gamma=0.8$ in \eqref{eqn:deadlock_resolve} and $\beta=50$ in \eqref{eqn:icbf}.

\subsection{$100\%$ collision guarantee in \ours}
\label{subsec: collision_avoidance}
As seen from Figs. \ref{fig:collision_empty} and \ref{fig:collision_avoidance}, our proposed approach (\ours) guarantees 100\% collision avoidance, regardless of the number of agents in both empty as well as maze environments. This scalability feature is particularly significant, as it allows \ours to handle scenarios with more than $>1000$ agents effectively.
Figs. \ref{fig:collision_empty} and ~\ref{fig:collision_avoidance} illustrate the performance of \ours as the number of agents increases, highlighting its ability to maintain collision-free trajectories in complex environments. From these experiments, we observe that in contrast to prior methods where the percentage of collision avoidance significantly decreases with the number of agents, \ours can provide safety guarantees even for a large number of agents. In addition, as seen from Fig.~\ref{fig:scalibility}, the computational load (runtime in ms) is not compromised as the number of agents increases, thereby showing good scalability. Our baselines, including \gcbf, \macbf, and MARL methods are purely learning based with soft safety guarantees (collision-checking) and prioritize goal completion at the cost of safety as the number of agents scale and have comparatively low safety.

\subsection{\textit{Minimizing deadlocks while generating safe trajectories: }}
\label{subsec: deadlock_prevention}
As observed from Figures \ref{fig:deadlock_empty} and \ref{fig:deadlock_maze}, there is a significant reduction in the number of deadlocks among agents in both empty and maze-like environments by leveraging the deadlock minimization mechanism described in \ref{deadlock_mechanism}.
For the experiments, the values of hyperparameters chosen were $\gamma = 0.99$, $\eta = 0.01$, and the threshold magnitude of control input for deadlock detection was taken as $0.01$. 
In addition, from Fig. \ref{fig:deadlock_maze}, one can observe that \ours reduces the average number of deadlocks by up to $\approx 92\%$ with $1024$ agents, compared to the baseline methods in a complex maze environment. Furthermore, 
the average number of deadlocks increases with the number of agents. However, the proposed method \ours successfully minimizes deadlocks while ensuring that the trajectories are safe. In addition, the safety and scalability of \ours are not compromised during deadlock minimization (see Figs. \ref{fig:collision_empty}, \ref{fig:collision_avoidance}, \ref{fig:scalibility_empty}, \ref{fig:scalibility}).

\vspace{0.5cm}

\subsection{\textit{Ensuring satisfaction of input constraints: }}
\label{subsec:input_constraints_satisfaction}
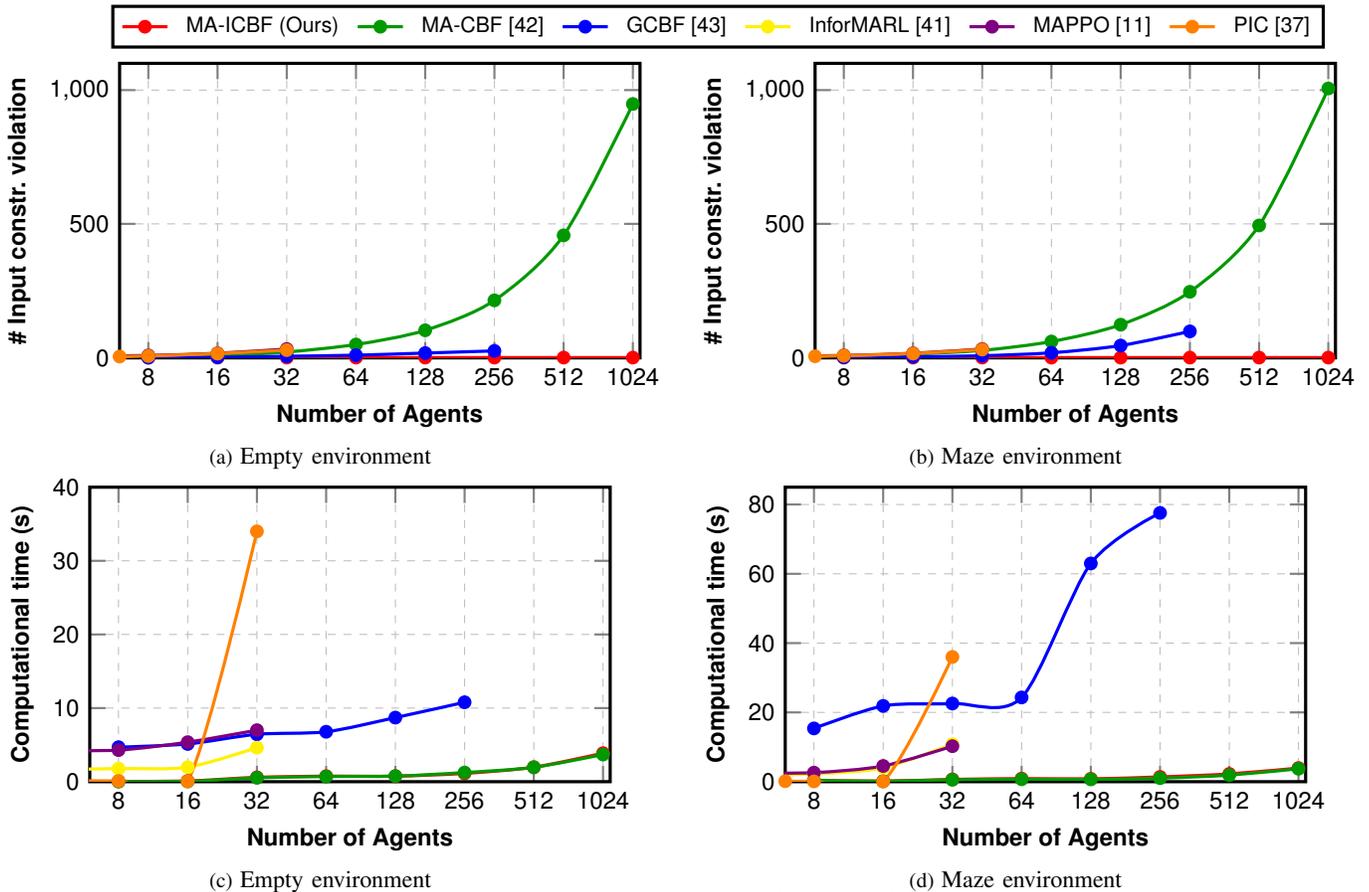
\begin{figure*}[]
    \centering
    \begin{subfigure}[b]{0.48\textwidth}
        \centering
            \captionsetup{skip=-5pt}\begin{tikzpicture}
\begin{axis}[
scale=\scaletikz,
    title style={font=\large\sansmath\sffamily, yshift=1ex},
    xlabel={\agentempty},
    ylabel={\textbf{\# Input constr. violation}},
    width=\widthtikz, 
    height=\heighttikz, 
    xtick={8, 16, 32, 64, 128, 256, 512, 1024},
    xticklabels={8, 16, 32, 64, 128, 256, 512, 1024},
    legend pos=north west,
    legend cell align={left},
    legend style={font=\small\sansmath\sffamily},
    xmajorgrids=true,
    ymajorgrids=true,
    grid style=dashed,
    font=\small\sansmath\sffamily,
    tick style={major tick length=6pt, minor tick length=3pt, thick},
    xmode=log,
    log basis x={2},
    every axis plot/.append style={thick},
    scaled ticks=false,
    xmin=6, xmax=1100,
    ymin=-1, ymax=1100,
    cycle list name=color list,
    line width=1.2pt,
        legend style={
        at={(1.15,1.05)},
        anchor=south,
    legend columns=6,    font=\small\sansmath\sffamily,
        column sep=1ex
    },
    legend cell align={left},
]

\addplot[
    color=red,
                    mark=*,
    mark options={solid, fill=red},
    line width=1.2pt,
    smooth
    ]
    coordinates {
        (8,0) (16,0) (32,0) (64,0) (128, 0) (256, 0) (512, 0) (1024, 0)
    };
\addlegendentry{\footnotesize \ours (Ours)}

\addplot[
    color=green!60!black,
                    mark=*,
    mark options={solid, fill=green!60!black},
    line width=1.2pt,
    smooth
    ]
    coordinates {
    (8,2) (16,7) (32,21) (64,49) (128, 102) (256, 214) (512, 457) (1024, 948)};
\addlegendentry{\footnotesize MA-CBF \cite{qin2021learning_macbf}}

\addplot[
    color=blue,
                    mark=*,
    mark options={solid, fill=blue},
    line width=1.2pt,
    smooth
    ]
    coordinates {
        (8,0) (16,1) (32,5) (64,9) (128, 17) (256, 25)
    };
\addlegendentry{\footnotesize GCBF \cite{zhang2023neural_gcbf}}

\addplot[
    color=yellow,
                    mark=*,
    mark options={solid, fill=yellow},
    line width=1.2pt,
    smooth
    ]
    coordinates {
        (4,4) (8,8) (16,16) (32,32) 
    };
\addlegendentry{\footnotesize InforMARL\cite{nayak22informarl}}

\addplot[
    color=violet,
                    mark=*,
    mark options={solid, fill=violet},
    line width=1.2pt,
    smooth
    ]
    coordinates {
        (4,4) (8,8) (16,16) (32,32) 
    };
\addlegendentry{\footnotesize MAPPO \cite{yu2022surprising_ma_lit_1}}

\addplot[
    color=orange,
                    mark=*,
    mark options={solid, fill=orange},
    line width=1.2pt,
    smooth
    ]
    coordinates {
        (6,4) (8,6) (16,15) (32,28) 
    };
\addlegendentry{\footnotesize PIC \cite{liu2020pic}}

\end{axis}
\end{tikzpicture}
        \caption{\small Empty environment}
\label{fig:input_constraint_satisfaction}
    \end{subfigure}
    \hfill
    \begin{subfigure}[b]{0.48\textwidth}
        \centering
            \captionsetup{skip=-5pt}
\vspace{-5em}
\begin{tikzpicture}
\begin{axis}[
scale=\scaletikz,
    title style={font=\large\sansmath\sffamily, yshift=-1.5ex},
    xlabel={\agentmaze},
    ylabel={\textbf{\# Input constr. violation}},
    width=\widthtikz, 
    height=\heighttikz, 
    xtick={8, 16, 32, 64, 128, 256, 512, 1024},
    xticklabels={8, 16, 32, 64, 128, 256, 512, 1024},
    legend pos=north west,
    legend cell align={left},
    legend style={font=\small\sansmath\sffamily},
    xmajorgrids=true,
    ymajorgrids=true,
    grid style=dashed,
    font=\small\sansmath\sffamily,
    tick style={major tick length=6pt, minor tick length=3pt, thick},
    xmode=log,
    log basis x={2},
    every axis plot/.append style={thick},
    scaled ticks=false,
    xmin=6, xmax=1100,
    ymin=-1, ymax=1100,
    cycle list name=color list,
    line width=1.2pt,
        legend style={
        at={(0.42,1.1)},
        anchor=south,
        legend columns=3,
        font=\small\sansmath\sffamily,
        column sep=1ex
    },
    legend cell align={left},
]

\addplot[
    color=red,
                    mark=*,
    mark options={solid, fill=red},
    line width=1.2pt,
    smooth
    ]
    coordinates {
        (8,0) (16,0) (32,0) (64,0) (128, 0) (256, 0) (512, 0) (1024, 0)
    };

\addplot[
    color=green!60!black,
                    mark=*,
    mark options={solid, fill=green!60!black},
    line width=1.2pt,
    smooth
    ]
    coordinates {
    (8,4) (16,13) (32,27) (64,60) (128, 123) (256, 246) (512, 494) (1024, 1006)};

\addplot[
    color=blue,
                    mark=*,
    mark options={solid, fill=blue},
    line width=1.2pt,
    smooth
    ]
    coordinates {
        (8,1) (16,3) (32,7) (64,18) (128, 45) (256, 98)
    };

\addplot[
    color=yellow,
                    mark=*,
    mark options={solid, fill=yellow},
    line width=1.2pt,
    smooth
    ]
    coordinates {
        (4,4) (8,8) (16,16) (32,32) 
    };

\addplot[
    color=violet,
                    mark=*,
    mark options={solid, fill=violet},
    line width=1.2pt,
    smooth
    ]
    coordinates {
        (4,4) (8,8) (16,16) (32,32) 
    };

\addplot[
    color=orange,
                    mark=*,
    mark options={solid, fill=orange},
    line width=1.2pt,
    smooth
    ]
    coordinates {
        (6,5) (8,7) (16,15) (32,30) 
    };

\end{axis}
\end{tikzpicture}
        \caption{\small Maze environment}
\label{fig:input_constraint_satisfaction_maze}
    \end{subfigure}
    \centering
    \begin{subfigure}[b]{0.48\textwidth}
        \centering
            \captionsetup{skip=-5pt}





\begin{tikzpicture}
\begin{axis}[
scale=\scaletikz,
    title style={font=\large\sansmath\sffamily, yshift=1ex},
    xlabel={\agentempty},
    ylabel={\textbf{Computational time (s)}},
    width=\widthtikz, 
    height=\heighttikz, 
    xtick={8, 16, 32, 64, 128, 256, 512, 1024},
    xticklabels={8, 16, 32, 64, 128, 256, 512, 1024},
    legend pos=north east,
    legend cell align={left},
    legend style={font=\small\sansmath\sffamily},
    xmajorgrids=true,
    ymajorgrids=true,
    grid style=dashed,
    font=\small\sansmath\sffamily,
    tick style={major tick length=6pt, minor tick length=3pt, thick},
    xmode=log,
    log basis x={2},
    every axis plot/.append style={thick},
    scaled ticks=false,
    xmin=6, xmax=1100,
    ymin=0, ymax=40.000,
    cycle list name=color list,
    line width=1.2pt,
        legend style={
        at={(1.3,1.1)},
        anchor=south,
        legend columns=6,
        font=\small\sansmath\sffamily,
        column sep=1ex
    },
    legend cell align={left},
]

\addplot[
    color=red,
                    mark=*,
    mark options={solid, fill=red},
    line width=1.2pt,
    smooth
    ]
    coordinates {
       (0, 0) (8,0.040) (16, 0.093) (32, 0.593) (64,0.750) (128, 0.750) (256, 1.128) (512,1.968) (1024,3.875)
    };

\addplot[
    color=green!60!black,
                    mark=*,
    mark options={solid, fill=green!60!black},
    line width=1.2pt,
    smooth
    ]
    coordinates {
       (0, 0) (8,0.027) (16, 0.044) (32,0.531) (64,0.700) (128,0.760) (256,1.240) (512,1.953) (1024,3.689)
    };

\addplot[
    color=blue,
                    mark=*,
    mark options={solid, fill=blue},
    line width=1.2pt,
    smooth
    ]
    coordinates {
       (0, 0) (8,4.685) (16,5.125) (32,6.437) (64, 6.781) (128, 8.703) (256, 10.786)
    };
\addplot[
    color=yellow,
                    mark=*,
    mark options={solid, fill=yellow},
    line width=1.2pt,
    smooth
    ]
    coordinates {
        (4,1.625) (8,1.780) (16,1.968) (32,4.625) 
    };

\addplot[
    color=violet,
                    mark=*,
    mark options={solid, fill=violet},
    line width=1.2pt,
    smooth
    ]
    coordinates {
        (4,4.250) (8,4.290) (16, 5.370) (32,7.000) 
    };

\addplot[
    color=orange,
                    mark=*,
    mark options={solid, fill=orange},
    line width=1.2pt,
    smooth
    ]
    coordinates {
        (4,0.155) (8, 0.100) (16,0.050) (32,34.000) 
    };

\end{axis}
\end{tikzpicture}
         \caption{\small Empty environment}
        \label{fig:scalibility_empty}
    \end{subfigure}
    \hfill
    \begin{subfigure}[b]{0.48\textwidth}
        \centering
            \captionsetup{skip=-5pt}





\begin{tikzpicture}
\begin{axis}[
scale=\scaletikz,
    title style={font=\large\sansmath\sffamily, yshift=1ex},
    xlabel={\agentmaze},
    ylabel={\textbf{Computational time (s)}},
    width=\widthtikz, 
    height=\heighttikz, 
    xtick={8, 16, 32, 64, 128, 256, 512, 1024},
    xticklabels={8, 16, 32, 64, 128, 256, 512, 1024},
    legend pos=north west,
    legend cell align={left},
    legend style={font=\small\sansmath\sffamily},
    legend style={
        font=\small\sansmath\sffamily,
        nodes={scale=0.75, transform shape},
    },
    xmajorgrids=true,
    ymajorgrids=true,
    grid style=dashed,
    font=\small\sansmath\sffamily,
    tick style={major tick length=6pt, minor tick length=3pt, thick},
    xmode=log,
    log basis x={2},
    every axis plot/.append style={thick},
    scaled ticks=false,
    xmin=6, xmax=1100,
    ymin=0, ymax=85.000,
    cycle list name=color list,
    line width=1.2pt,
        legend style={
        at={(0.52,1.1)},
        anchor=south,
        legend columns=3,
        font=\small\sansmath\sffamily,
        column sep=1ex
    },
    legend cell align={left},
]

\addplot[
    color=red,
                    mark=*,
    mark options={solid, fill=red},
    line width=1.2pt,
    smooth
    ]
    coordinates {
       (0, 0) (8,0.400) (16, 0.160) (32, 0.660) (64,0.890) (128, 0.830) (256, 1.380) (512,2.240) (1024,3.940)
    };

\addplot[
    color=green!60!black,
                    mark=*,
    mark options={solid, fill=green!60!black},
    line width=1.2pt,
    smooth
    ]
    coordinates {
       (0, 0) (8,0.370) (16,0.140) (32,0.560) (64,0.703) (128,0.671) (256,1.015) (512,1.880) (1024,3.740)
    };

\addplot[
    color=blue,
                    mark=*,
    mark options={solid, fill=blue},
    line width=1.2pt,
    smooth
    ]
    coordinates {
       (0, 0) (8,15.375) (16,21.870) (32,22.560) (64,24.310) (128, 62.984) (256, 77.570)
    };

\addplot[
    color=yellow,
                    mark=*,
    mark options={solid, fill=yellow},
    line width=1.2pt,
    smooth
    ]
    coordinates {
        (4,1.940) (8,2.250) (16,4.030) (32,10.830) 
    };

\addplot[
    color=violet,
                    mark=*,
    mark options={solid, fill=violet},
    line width=1.2pt,
    smooth
    ]
    coordinates {
        (4,2.500) (8,2.620) (16,4.540) (32,10.240) 
    };

\addplot[
    color=orange,
                    mark=*,
    mark options={solid, fill=orange},
    line width=1.2pt,
    smooth
    ]
    coordinates {
        (6,0.145) (8, 0.080) (16,0.070) (32,36.000) 
    };

\end{axis}
\end{tikzpicture}
         \caption{\small Maze environment}
        \label{fig:scalibility}
    \end{subfigure}
    \caption{\small(a)-(b): Average number of input constraint violations versus the number of agents in different settings. (c)-(d) Average computational time per agent in milliseconds in different environments}
    \label{fig:scalibility_envs}
    \vspace{-10pt}
\end{figure*}

We set the input constraints in the range of $[-0.2, 0.2]$ meters per second for the linear velocity and $[-12, 12]$ degrees for the angular velocity. Figs. \ref{fig:input_constraint_satisfaction} and \ref{fig:input_constraint_satisfaction_maze} show that the proposed approach \ours successfully constrains all agents within these bounds. In contrast, baselines \gcbf, \macbf, and MARL methods show an exponential increase in the number of input constraint violations
with a maximum of $1006$ agents (out of $1024$ agents) which means that $\approx 98\%$ agents have violated the input constraints. 

\subsection{\textit{Computational benefits using log-sum-exp trick \eqref{eqn:icbf} in MPC-ICBF: }}
\label{subsec: log-sum}
\begin{figure}[H]
    \centering

\begin{tikzpicture}
    \begin{axis}[
     scale=0.75,
        ybar,
        bar width=0.5cm,
        enlarge x limits={abs=1cm},
        symbolic x coords={64, 128, 256, 512, 1024},
        xtick=data,
        xlabel={ \textbf{Number of agents (in maze env.)}},
        ylabel={ \textbf{Avg. comp. time (s)}},
        ymin=0, ymax=4.600,
        nodes near coords,
        nodes near coords align={vertical},
        every node near coord/.append style={font=\footnotesize},
        xticklabel style={font=\small},
        yticklabel style={font=\small},
        xlabel style={font=\small},
        ylabel style={font=\small},
        title style={font=\small, yshift=1ex},
        legend style={at={(0.6,0.93)}, anchor=north east, legend columns=1, font=\large, draw=none},
        grid=both,
        minor tick num=1,
        major grid style={line width=.2pt,draw=gray!50},
        minor grid style={line width=.1pt,draw=gray!20},
    ]
        \addplot[draw=none,fill=orange!80!black] coordinates { (256, 1.503) (512, 2.490) (1024, 4.140)};
        \addplot[draw=none,fill=red!70!blue] coordinates {(256, 0.988) (512, 1.240) (1024, 2.100)};
        \legend{\footnotesize NN+MPC, \footnotesize \ours (Ours)}
    \end{axis}
\end{tikzpicture}
    \caption{\small The average computational time in seconds per agent of trivially encoding MPC and NN based controller versus \ours}
    \centering
    \label{fig:scale_mpc}
    \vspace{-10pt}
\end{figure}
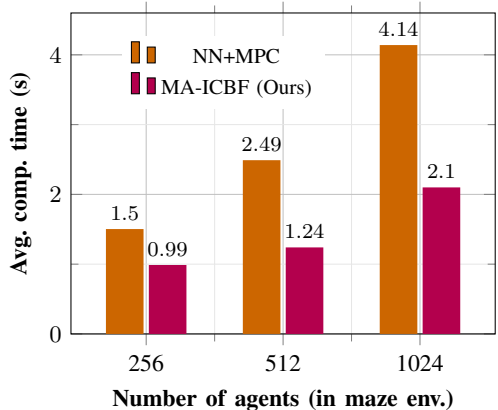

As discussed in the ~\ref{subsec:collision_avoidance_and_deadlock_analysis}, the log-sum-exp trick converts the multiple collision avoidance constraints \eqref{eqn:mpc_trivial} into a single constraint \eqref{eqn:mpc_proposed_via_icbf} for an agent. 
Furthermore, to accelerate the summation in the log-sum-exp expression in \eqref{eqn:icbf}, we leverage parallelization using the \texttt{cupy}~\cite{nishino2017cupy} package. We compare this approach with a baseline that trivially augments the learned NN control policy with the MPC based controller with ICBF constraints (NN+MPC). Fig.~\ref{fig:scale_mpc} shows the histogram plot comparing the computational time per agent in milliseconds for two methods— \ours using the log-sum-exp trick and the baseline method (NN+MPC). As observed from Fig. \ref{fig:scale_mpc}, there is a significant computational advantage ($\approx$ $1.5-2\operatorname{x}$) between these two approaches. Furthermore, with the increase in the number of agents, the computational benefits become more apparent. This is mainly because, with the higher number of agents, the number of collisions increases, making the log-sum-exp trick's ability to reduce the computational load more noticeable.
\subsection{\textit{Generalizibility: }}
     \begin{table}[]
 \scriptsize\centering
        \begin{tabular}{lcccc}
        \toprule
        \multicolumn{2}{c}{\diagbox{\bf Train}{\bf Test}}& $\boldsymbol{n=256}$  & $\boldsymbol{n=512}$ & $\boldsymbol{n=1024}$\\ 
        \midrule
            \multirow{2}{*}{$\boldsymbol{m=4}$} 
            & $\#$MPC-ICBF trig.  & 21  & 73 & 280 \\ 
            & $\# \text{Deadlock}$ & 6 & 15 & 18\\ 
             \midrule
            
            \multirow{2}{*}{$\boldsymbol{m=8}$} 
            & $\#$MPC-ICBF trig. & 0  & 1  & 10 \\ 
            & $\# \text{Deadlock}$ & 0 & 0 & 0\\ 
              \midrule
            
            \multirow{2}{*}{$\boldsymbol{m=16}$} 
            & $\#$MPC-ICBF trig. & 0 & 0 & 2\\ 
            & $\# \text{Deadlock}$ & 0 & 0 & 0\\ 
            \bottomrule
        \end{tabular}
       \vspace{5pt}
        \caption{\small Test performance of \ours for the \emph{maze} environment, when trained with $m$ agents and tested with $n$ agents.
        }
        \label{table:gen}
        \vspace{-15pt}
    \end{table}
Table ~\ref{table:gen}, shows how generalizable \ours is, when trained with fewer agents ($4$, $8$ or $16$) and then tested with a larger number of agents ($256$, $512$, or $1024$) in a maze environment. From table \ref{table:gen}, when trained with $16$ agents, \ours performs well even with $1024$ agents with no deadlocks, while at the same triggering the MPC-ICBF module twice to ensure complete collision avoidance and input constraint satisfaction.

\section{Conclusion\label{sec:conclusion}} 
In this paper, we have addressed the problem of multi-agent control under limited actuation, particularly focusing on the aspects of scalability, safety guarantees, and input constraint satisfaction. Our proposed approach first jointly learns a decentralized neural Integral Control Barrier (ICBF) certificate and a decentralized safe neural network (NN) controller. During the execution phase, the safe learned controller is augmented with a computationally light MPC-ICBF based framework as well as a deadlock minimization module to guarantee safety and minimize deadlocks respectively. The main limitation of our work is the assumption that the governing dynamics of each agent are known a priori. Future work includes extending this work to vision-control tasks, particularly focused on multi-agent control that is also able to handle dynamic obstacles. 

\bibliography{main.bib}

\begin{thebibliography}{10}
\providecommand{\url}[1]{#1}
\csname url@samestyle\endcsname
\providecommand{\newblock}{\relax}
\providecommand{\bibinfo}[2]{#2}
\providecommand{\BIBentrySTDinterwordspacing}{\spaceskip=0pt\relax}
\providecommand{\BIBentryALTinterwordstretchfactor}{4}
\providecommand{\BIBentryALTinterwordspacing}{\spaceskip=\fontdimen2\font plus
\BIBentryALTinterwordstretchfactor\fontdimen3\font minus \fontdimen4\font\relax}
\providecommand{\BIBforeignlanguage}[2]{{%
\expandafter\ifx\csname l@#1\endcsname\relax
\typeout{** WARNING: IEEEtran.bst: No hyphenation pattern has been}%
\typeout{** loaded for the language `#1'. Using the pattern for}%
\typeout{** the default language instead.}%
\else
\language=\csname l@#1\endcsname
\fi
#2}}
\providecommand{\BIBdecl}{\relax}
\BIBdecl

\bibitem{li2023motion_warehouse_automation}
X.~Li, ``Motion planning of warehouse automation robot based on recurrent neural network,'' in \emph{2023 International Conference on Mechatronics, IoT and Industrial Informatics (ICMIII)}.\hskip 1em plus 0.5em minus 0.4em\relax IEEE, 2023, pp. 385--389.

\bibitem{claussmann2019review_autonomous_driving}
L.~Claussmann, M.~Revilloud, D.~Gruyer, and S.~Glaser, ``A review of motion planning for highway autonomous driving,'' \emph{IEEE Transactions on Intelligent Transportation Systems}, vol.~21, no.~5, pp. 1826--1848, 2019.

\bibitem{batra2022decentralized_quad_swarm_1}
S.~Batra, Z.~Huang, A.~Petrenko, T.~Kumar, A.~Molchanov, and G.~S. Sukhatme, ``Decentralized control of quadrotor swarms with end-to-end deep reinforcement learning,'' in \emph{Conference on Robot Learning}.\hskip 1em plus 0.5em minus 0.4em\relax PMLR, 2022, pp. 576--586.

\bibitem{park2022online_quad_swarm_2}
J.~Park, D.~Kim, G.~C. Kim, D.~Oh, and H.~J. Kim, ``Online distributed trajectory planning for quadrotor swarm with feasibility guarantee using linear safe corridor,'' \emph{IEEE Robotics and Automation Letters}, vol.~7, no.~2, pp. 4869--4876, 2022.

\bibitem{au2010motion_intersection_management}
T.-C. Au and P.~Stone, ``Motion planning algorithms for autonomous intersection management,'' in \emph{Workshops at the Twenty-Fourth AAAI Conference on Artificial Intelligence}, 2010.

\bibitem{sharon2015conflict_classical_cbs_1}
G.~Sharon, R.~Stern, A.~Felner, and N.~R. Sturtevant, ``Conflict-based search for optimal multi-agent pathfinding,'' \emph{Artificial intelligence}, vol. 219, pp. 40--66, 2015.

\bibitem{kottinger2022conflict_classical_cbs_2}
J.~Kottinger, S.~Almagor, and M.~Lahijanian, ``Conflict-based search for explainable multi-agent path finding,'' in \emph{Proceedings of the International Conference on Automated Planning and Scheduling}, vol.~32, 2022, pp. 692--700.

\bibitem{dergachev2021distributed_classical_3}
S.~Dergachev and K.~Yakovlev, ``Distributed multi-agent navigation based on reciprocal collision avoidance and locally confined multi-agent path finding,'' in \emph{2021 IEEE 17th International Conference on Automation Science and Engineering (CASE)}.\hskip 1em plus 0.5em minus 0.4em\relax IEEE, 2021, pp. 1489--1494.

\bibitem{alonso2013optimal_classical_4}
J.~Alonso-Mora, A.~Breitenmoser, M.~Rufli, P.~Beardsley, and R.~Siegwart, ``Optimal reciprocal collision avoidance for multiple non-holonomic robots,'' in \emph{Distributed autonomous robotic systems: The 10th international symposium}.\hskip 1em plus 0.5em minus 0.4em\relax Springer, 2013, pp. 203--216.

\bibitem{chandra2023decentralized}
R.~Chandra, V.~Zinage, E.~Bakolas, J.~Biswas, and P.~Stone, ``Decentralized multi-robot social navigation in constrained environments via game-theoretic control barrier functions,'' \emph{arXiv preprint arXiv:2308.10966}, 2023.

\bibitem{yu2022surprising_ma_lit_1}
C.~Yu, A.~Velu, E.~Vinitsky, J.~Gao, Y.~Wang, A.~Bayen, and Y.~Wu, ``The surprising effectiveness of ppo in cooperative multi-agent games,'' \emph{Advances in Neural Information Processing Systems}, vol.~35, pp. 24\,611--24\,624, 2022.

\bibitem{ames2019control_ma_lit_2}
A.~D. Ames, S.~Coogan, M.~Egerstedt, G.~Notomista, K.~Sreenath, and P.~Tabuada, ``Control barrier functions: Theory and applications,'' in \emph{2019 18th European control conference (ECC)}.\hskip 1em plus 0.5em minus 0.4em\relax IEEE, 2019, pp. 3420--3431.

\bibitem{glotfelter2017nonsmooth_ma_lit_3}
P.~Glotfelter, J.~Cort{\'e}s, and M.~Egerstedt, ``Nonsmooth barrier functions with applications to multi-robot systems,'' \emph{IEEE control systems letters}, vol.~1, no.~2, pp. 310--315, 2017.

\bibitem{jankovic2021collision_ma_lit_4}
M.~Jankovic and M.~Santillo, ``Collision avoidance and liveness of multi-agent systems with cbf-based controllers,'' in \emph{2021 60th IEEE Conference on Decision and Control (CDC)}.\hskip 1em plus 0.5em minus 0.4em\relax IEEE, 2021, pp. 6822--6828.

\bibitem{cheng2020safe_ma_lit_5}
R.~Cheng, M.~J. Khojasteh, A.~D. Ames, and J.~W. Burdick, ``Safe multi-agent interaction through robust control barrier functions with learned uncertainties,'' in \emph{2020 59th IEEE Conference on Decision and Control (CDC)}.\hskip 1em plus 0.5em minus 0.4em\relax IEEE, 2020, pp. 777--783.

\bibitem{garg2021robust_ma_lit_6}
K.~Garg and D.~Panagou, ``Robust control barrier and control lyapunov functions with fixed-time convergence guarantees,'' in \emph{2021 American Control Conference (ACC)}.\hskip 1em plus 0.5em minus 0.4em\relax IEEE, 2021, pp. 2292--2297.

\bibitem{kohler2024distributed_opt_1}
M.~K{\"o}hler, M.~A. M{\"u}ller, and F.~Allg{\"o}wer, ``Distributed mpc for self-organized cooperation of multi-agent systems,'' \emph{IEEE Transactions on Automatic Control}, 2024.

\bibitem{saccani2023model_opt_2}
D.~Saccani, L.~Fagiano, M.~N. Zeilinger, and A.~Carron, ``Model predictive control for multi-agent systems under limited communication and time-varying network topology,'' in \emph{2023 62nd IEEE Conference on Decision and Control (CDC)}.\hskip 1em plus 0.5em minus 0.4em\relax IEEE, 2023, pp. 3764--3769.

\bibitem{chen2020guaranteed_opt_3}
Y.~Chen, A.~Singletary, and A.~D. Ames, ``Guaranteed obstacle avoidance for multi-robot operations with limited actuation: A control barrier function approach,'' \emph{IEEE Control Systems Letters}, vol.~5, no.~1, pp. 127--132, 2020.

\bibitem{wang2017safety_opt_4}
L.~Wang, A.~D. Ames, and M.~Egerstedt, ``Safety barrier certificates for collisions-free multirobot systems,'' \emph{IEEE Transactions on Robotics}, vol.~33, no.~3, pp. 661--674, 2017.

\bibitem{xiao2019control_cbf_high_1}
W.~Xiao and C.~Belta, ``Control barrier functions for systems with high relative degree,'' in \emph{2019 IEEE 58th conference on decision and control (CDC)}.\hskip 1em plus 0.5em minus 0.4em\relax IEEE, 2019, pp. 474--479.

\bibitem{wang2021learning_cbf_high_2}
C.~Wang, Y.~Meng, Y.~Li, S.~L. Smith, and J.~Liu, ``Learning control barrier functions with high relative degree for safety-critical control,'' in \emph{2021 European Control Conference (ECC)}.\hskip 1em plus 0.5em minus 0.4em\relax IEEE, 2021, pp. 1459--1464.

\bibitem{marley2024hybrid_1}
M.~Marley, R.~Skjetne, and A.~R. Teel, ``Hybrid control barrier functions for continuous-time systems,'' \emph{IEEE Transactions on Automatic Control}, 2024.

\bibitem{lindemann2021learning_hybrid_2}
L.~Lindemann, H.~Hu, A.~Robey, H.~Zhang, D.~Dimarogonas, S.~Tu, and N.~Matni, ``Learning hybrid control barrier functions from data,'' in \emph{Conference on Robot Learning}.\hskip 1em plus 0.5em minus 0.4em\relax PMLR, 2021, pp. 1351--1370.

\bibitem{taylor2022safety_sampled_1}
A.~J. Taylor, V.~D. Dorobantu, R.~K. Cosner, Y.~Yue, and A.~D. Ames, ``Safety of sampled-data systems with control barrier functions via approximate discrete time models,'' in \emph{2022 IEEE 61st Conference on Decision and Control (CDC)}.\hskip 1em plus 0.5em minus 0.4em\relax IEEE, 2022, pp. 7127--7134.

\bibitem{niu2021safety_sampled_2}
L.~Niu, H.~Zhang, and A.~Clark, ``Safety-critical control synthesis for unknown sampled-data systems via control barrier functions,'' in \emph{2021 60th IEEE Conference on Decision and Control (CDC)}.\hskip 1em plus 0.5em minus 0.4em\relax IEEE, 2021, pp. 6806--6813.

\bibitem{zinage2023neural_unknown_zinage_1}
V.~Zinage and E.~Bakolas, ``Neural koopman control barrier functions for safety-critical control of unknown nonlinear systems,'' in \emph{2023 American Control Conference (ACC)}.\hskip 1em plus 0.5em minus 0.4em\relax IEEE, 2023, pp. 3442--3447.

\bibitem{zinage2023neural_icbf}
V.~Zinage, R.~Chandra, and E.~Bakolas, ``Neural differentiable integral control barrier functions for unknown nonlinear systems with input constraints,'' \emph{arXiv preprint arXiv:2312.07345}, 2023.

\bibitem{zinage2023neuralunknown}
V.~\textsc{Z}inage and E.~Bakolas, ``Neural koopman lyapunov control,'' \emph{Neurocomputing}, vol. 527, pp. 174--183, 2023.

\bibitem{jankovic2018control_input_delay_systems}
M.~Jankovic, ``Control barrier functions for constrained control of linear systems with input delay,'' in \emph{2018 annual American control conference (ACC)}.\hskip 1em plus 0.5em minus 0.4em\relax IEEE, 2018, pp. 3316--3321.

\bibitem{ames2020integral_ames}
A.~D. Ames, G.~Notomista, Y.~Wardi, and M.~Egerstedt, ``Integral control barrier functions for dynamically defined control laws,'' \emph{IEEE control systems letters}, vol.~5, no.~3, pp. 887--892, 2020.

\bibitem{zinage2023disturbance_integral_zinage}
V.~Zinage, R.~Chandra, and E.~Bakolas, ``Disturbance observer-based robust integral control barrier functions for nonlinear systems with high relative degree,'' \emph{arXiv preprint arXiv:2309.16945}, 2023.

\bibitem{long2018towards_learning_lit_1}
P.~Long, T.~Fan, X.~Liao, W.~Liu, H.~Zhang, and J.~Pan, ``Towards optimally decentralized multi-robot collision avoidance via deep reinforcement learning,'' in \emph{2018 IEEE international conference on robotics and automation (ICRA)}.\hskip 1em plus 0.5em minus 0.4em\relax IEEE, 2018, pp. 6252--6259.

\bibitem{chen2017decentralized_learning_lit_2}
Y.~F. Chen, M.~Liu, M.~Everett, and J.~P. How, ``Decentralized non-communicating multiagent collision avoidance with deep reinforcement learning,'' in \emph{2017 IEEE international conference on robotics and automation (ICRA)}.\hskip 1em plus 0.5em minus 0.4em\relax IEEE, 2017, pp. 285--292.

\bibitem{everett2021collision_learning_lit_3}
M.~Everett, Y.~F. Chen, and J.~P. How, ``Collision avoidance in pedestrian-rich environments with deep reinforcement learning,'' \emph{Ieee Access}, vol.~9, pp. 10\,357--10\,377, 2021.

\bibitem{kamenev2022predictionnet_learning_lit_4}
A.~Kamenev, L.~Wang, O.~B. Bohan, I.~Kulkarni, B.~Kartal, A.~Molchanov, S.~Birchfield, D.~Nist{\'e}r, and N.~Smolyanskiy, ``Predictionnet: Real-time joint probabilistic traffic prediction for planning, control, and simulation,'' in \emph{2022 International Conference on Robotics and Automation (ICRA)}.\hskip 1em plus 0.5em minus 0.4em\relax IEEE, 2022, pp. 8936--8942.

\bibitem{liu2020pic}
I.-J. Liu, R.~A. Yeh, and A.~G. Schwing, ``Pic: permutation invariant critic for multi-agent deep reinforcement learning,'' in \emph{Conference on Robot Learning}.\hskip 1em plus 0.5em minus 0.4em\relax PMLR, 2020, pp. 590--602.

\bibitem{zhang2021multi_marl_review}
K.~Zhang, Z.~Yang, and T.~Ba{\c{s}}ar, ``Multi-agent reinforcement learning: A selective overview of theories and algorithms,'' \emph{Handbook of reinforcement learning and control}, pp. 321--384, 2021.

\bibitem{wang2022model_based_marl}
X.~Wang, Z.~Zhang, and W.~Zhang, ``Model-based multi-agent reinforcement learning: Recent progress and prospects,'' \emph{arXiv preprint arXiv:2203.10603}, 2022.

\bibitem{yu2022surprising_mappo}
C.~Yu, A.~Velu, E.~Vinitsky, J.~Gao, Y.~Wang, A.~Bayen, and Y.~Wu, ``The surprising effectiveness of ppo in cooperative multi-agent games,'' \emph{Advances in Neural Information Processing Systems}, vol.~35, pp. 24\,611--24\,624, 2022.

\bibitem{nayak22informarl}
S.~Nayak, K.~Choi, W.~Ding, S.~Dolan, K.~Gopalakrishnan, and H.~Balakrishnan, ``Scalable multi-agent reinforcement learning through intelligent information aggregation,'' in \emph{International Conference on Machine Learning}.\hskip 1em plus 0.5em minus 0.4em\relax PMLR, 2023, pp. 25\,817--25\,833.

\bibitem{qin2021learning_macbf}
Z.~Qin, K.~Zhang, Y.~Chen, J.~Chen, and C.~Fan, ``Learning safe multi-agent control with decentralized neural barrier certificates,'' \emph{arXiv preprint arXiv:2101.05436}, 2021.

\bibitem{zhang2023neural_gcbf}
S.~Zhang, K.~Garg, and C.~Fan, ``Neural graph control barrier functions guided distributed collision-avoidance multi-agent control,'' in \emph{Conference on robot learning}.\hskip 1em plus 0.5em minus 0.4em\relax PMLR, 2023, pp. 2373--2392.

\bibitem{tee2009barrier_cbf_1}
K.~P. Tee, S.~S. Ge, and E.~H. Tay, ``Barrier lyapunov functions for the control of output-constrained nonlinear systems,'' \emph{Automatica}, vol.~45, no.~4, pp. 918--927, 2009.

\bibitem{qi2017pointnet}
C.~R. Qi, H.~Su, K.~Mo, and L.~J. Guibas, ``Pointnet: Deep learning on point sets for 3d classification and segmentation,'' in \emph{Proceedings of the IEEE conference on computer vision and pattern recognition}, 2017, pp. 652--660.

\bibitem{nishino2017cupy}
R.~Nishino and S.~H.~C. Loomis, ``Cupy: A numpy-compatible library for nvidia gpu calculations,'' \emph{31st confernce on neural information processing systems}, vol. 151, no.~7, 2017.

\end{thebibliography}

\end{document}